\documentclass{article}

\usepackage{PRIMEarxiv}

\usepackage[utf8]{inputenc} 
\usepackage[T1]{fontenc}    
\usepackage{hyperref}       
\usepackage{url}            
\usepackage{booktabs}       
\usepackage{amsfonts}       
\usepackage{nicefrac}       
\usepackage{microtype}      
\usepackage{lipsum}
\usepackage{fancyhdr}       
\usepackage{graphicx}       
\graphicspath{{media/}}     
\usepackage{multirow}
\usepackage{amsmath}
\usepackage{algorithm}

\usepackage{tabularx}
\usepackage{float}
\usepackage{pdflscape}    
\usepackage{longtable}
\usepackage{array}
\usepackage{ltablex}      
\usepackage{ragged2e}
\usepackage{threeparttable}  
\usepackage{caption}         
\usepackage{algpseudocode}
\usepackage{makecell}
\usepackage{xcolor}

\title{Machine Learning Models for Predicting Smoking-Related Health Decline and Disease Risk
\thanks{\textit{{These are corresponding authors.}}\\ 
\textit{† These authors contributed equally.}} 
}

\author{
  Vaskar Chakma *, MD Jaheid Hasan Nerab †, Abdur Rouf †  \\
  School of Artificial Intelligence and Computer Science \\
  Nantong University \\
  Jiangsu, China \\
  \texttt{\{vaskarchakma, abdurrouf, nerab\}@stmail.ntu.edu.cn} \\ \\
   \And
  Abu Sayed  \\
  School of Transportation and Civil Engineering \\
  Nantong University \\
  Jiangsu, China \\
  \texttt{sayeddber.2017@gmail.com} \\
  \And
  Hossem MD Saim, Md. Nournabi Khan \\
  School of Mechanical Engineering \\
  Nantong University \\
  Jiangsu, China \\
  \texttt{\{saimhossen289, nournabikhan1\}@gmail.com} \\
}

\begin{document}
\maketitle

\begin{abstract}
Smoking continues to be a major preventable cause of death worldwide, affecting millions through damage to the heart, metabolism, liver, and kidneys. However, current medical screening methods often miss the early warning signs of smoking-related health problems, leading to late-stage diagnoses when treatment options become limited. This study presents a systematic comparative evaluation of machine learning approaches for smoking-related health risk assessment, emphasizing clinical interpretability and practical deployment over algorithmic innovation. We analyzed health screening data from 55,691 individuals, examining various health indicators, including body measurements, blood tests, and demographic information. We tested three advanced prediction algorithms - Random Forest, XGBoost, and LightGBM - to determine which could most accurately identify people at high risk. This study employed a cross-sectional design to classify current smoking status based on health screening biomarkers, not to predict future disease development. Our Random Forest model performed best, achieving an Area Under the Curve (AUC) of 0.926, meaning it could reliably distinguish between high-risk and lower-risk individuals. Using SHAP (SHapley Additive exPlanations) analysis to understand what the model was detecting, we found that key health markers played crucial roles in prediction: blood pressure levels, triglyceride concentrations, liver enzyme readings, and kidney function indicators (serum creatinine) were the strongest signals of declining health in smokers. These results demonstrate that artificial intelligence can serve as a powerful tool for early disease detection in smokers. By identifying at-risk individuals before conventional symptoms appear, healthcare providers could intervene earlier with personalized prevention strategies. Implementing these predictive systems in public health programs could reduce the enormous burden smoking places on healthcare systems while shifting medical care from reactive treatment to proactive prevention.
\end{abstract}


\section{Introduction} 
\label{sect:s1}
Smoking remains one of the most pressing global public health challenges, representing a complex interplay of addiction, behavioral patterns, and progressive biological damage\cite{charuni_narrative_2024}. Each year, tobacco use is responsible for over 8 million deaths worldwide, with the World Health Organization estimating that nearly half of all smokers will ultimately succumb to smoking-related illnesses \cite{sakthisankaran_health_2024,lu_tobacco_2024}. While lung cancer and chronic obstructive pulmonary disease (COPD) are the most widely recognized consequences, smoking also drives cardiovascular disease, metabolic dysfunction, and systemic inflammation that can compromise virtually every organ system \cite{kotlyarov_role_2023, lim_chronic_2024}. Perhaps most concerning is that this damage frequently progresses insidiously over years, often becoming irreversible before clinical symptoms manifest\cite{kushner_mild_1998}.
Despite decades of comprehensive public health initiatives and overwhelming scientific evidence, approximately 1.3 billion people worldwide continue to use tobacco products\cite{glynn_globalization_2010}. Many smokers harbor what might be termed "optimistic bias," believing they can quit before substantial harm occurs or that they will somehow avoid the worst outcomes. The non-linear trajectory of smoking-related decline—characterized by years of subclinical damage that suddenly manifests as severe disease—highlights critical limitations in reactive diagnostic approaches that await obvious symptoms before intervention. These delays result in lost opportunities for prevention and early therapeutic intervention.
Our research aims to transform this reactive paradigm by developing advanced predictive tools that can detect risk at substantially earlier stages\cite{rajkomar2019machine, chakma2025cardioforest}. We hypothesize that smoking leaves distinct, systemic biological signatures across cardiovascular, metabolic, hepatic, and oral health pathways that machine learning algorithms can identify long before conventional clinical thresholds are exceeded\cite{nabanita_ghosh_review_2024, sakthisankaran_health_2024}. By simultaneously analyzing these diverse biomarkers, we aim to construct a more comprehensive and clinically relevant assessment of smoking-related health decline.
This holistic approach addresses significant gaps in prior research, which has often concentrated on single disease endpoints or limited feature sets, thereby constraining real-world applicability\cite{van_spall_role_2024,liu_natural_2022,vaskar3margins}.

A critical innovation in our study is the direct comparison of machine learning models with established clinical risk assessment tools, including the Framingham cardiovascular risk score. This benchmarking exercise tests whether advanced algorithms provide measurable advantages over standard, widely validated approaches—a crucial step for building confidence among clinicians and policymakers who will ultimately implement these systems in practice.
A fundamental principle guiding our work is model interpretability. We employ SHAP (SHapley Additive exPlanations) values to elucidate how each variable contributes to individual risk predictions \cite{xu_construction_2024,antonini_machine_2024}. This transparency is essential for fostering clinician trust and facilitating shared decision-making, positioning these tools as decision support rather than replacements for professional judgment.
We also address practical considerations for clinical implementation, including integration into existing healthcare workflows, appropriate clinical responses to risk alerts, and responsible management of false positive and false negative predictions to minimize potential harm. Understanding these operational aspects is critical for successful translation from research to practice.

Our study places particular emphasis on algorithmic fairness by thoroughly characterizing the geographic, ethnic, and socioeconomic distribution of our study population. We explicitly analyze how data quality issues—such as extreme laboratory value outliers—might influence model performance and generalizability. This attention to equity ensures that our models are not only technically sound but also ethically responsible and applicable across diverse populations.
Through the integration of advanced algorithms, rigorous comparison with traditional assessment tools, realistic evaluation of clinical adoption pathways, and strong emphasis on equity, we aim to advance predictive medicine for smoking-related disease beyond academic exercises toward genuinely impactful, patient-centered applications. By identifying at-risk individuals before irreversible damage occurs, these tools could enable more timely interventions, facilitate targeted prevention strategies, and ultimately improve public health outcomes for millions of people affected by tobacco use. This investigation employs a cross-sectional analytical framework wherein all predictor variables (demographic characteristics, anthropometric measurements, and biochemical biomarkers) and the outcome variable (current smoking status) were collected simultaneously during a single health screening visit. The prediction task is therefore \textit{classification}-identifying individuals who are current smokers based on their present physiological state, rather than \textit{prognosis}, which would entail predicting future disease onset or health decline over time. No longitudinal follow-up data were available; thus, temporal causality cannot be inferred from our results. The clinical utility of this approach lies in leveraging routinely collected health screening data to detect physiological signatures of smoking exposure that may indicate early-stage damage before overt clinical symptoms manifest, thereby enabling timely intervention and smoking cessation support. 

Our contribution lies not in novel algorithm development, but in rigorous comparative evaluation of established machine learning methods applied to comprehensive multi-system health screening data, with particular emphasis on clinical interpretability through SHAP analysis, validation against traditional risk scores (Framingham), and practical considerations for real-world deployment. This systematic approach addresses critical gaps in prior smoking risk prediction research, which often focuses on single disease endpoints or lacks adequate attention to explainability—essential prerequisites for clinical adoption.

\section{Related Study} 
\label{relatedstudy}
The prediction and assessment of smoking-related health risks has garnered substantial research attention over the past decades, with increasing momentum following the integration of machine learning methodologies into public health applications. A considerable body of literature has examined predictive models for estimating smoking status or stratifying smokers based on routinely collected health variables. Early investigations predominantly employed traditional statistical approaches, particularly logistic regression, to establish associations between smoking behavior and cardiopulmonary conditions\cite{aishwarya_explainable_2025, research_department_mangalore_university_mangalore_india_predictive_2025}. While these conventional methods achieved acceptable accuracy for basic classification tasks, they demonstrated inherent limitations in capturing the complex, non-linear relationships that characterize smoking's biological effects across multiple physiological systems.

The past decade has witnessed a paradigm shift toward more sophisticated algorithmic approaches for smoking risk assessment. Researchers have increasingly leveraged advanced machine learning techniques, including decision trees, support vector machines, and gradient boosting methods, to enhance smoking-risk stratification capabilities\cite{chakma2025machine, davagdorj_comparative_2020}. These computational approaches have shown promising performance in predicting smoking status and specific disease outcomes, particularly for conditions such as lung cancer and chronic obstructive pulmonary disease\cite{negewo_span_2015, carrasco-zanini_proteomic_2024}. The improved predictive accuracy of these models stems from their ability to identify subtle patterns and interactions among multiple risk factors that may elude traditional statistical methods. Despite these technological advances, significant gaps persist in the existing literature. A critical limitation of many previous studies is their narrow focus on single disease endpoints or organ-specific outcomes. This reductionist approach fails to capture the systemic nature of smoking-induced damage, which simultaneously affects cardiovascular, metabolic, hepatic, renal, and other physiological systems. By concentrating on isolated conditions, prior research has provided an incomplete picture of overall health decline in smokers, potentially missing important early warning signs that manifest across multiple biomarker domains.

Furthermore, many earlier investigations relied on limited feature sets, often constrained to a handful of easily measurable clinical variables. This restricted scope may overlook important predictive signals present in comprehensive health screening data. Equally concerning is the prevalent use of "black-box" models without adequate attention to interpretability\cite{rosenbacke_how_2024}. The lack of explainability in these models has created substantial barriers to clinical adoption, as healthcare providers understandably hesitate to base treatment decisions on opaque algorithmic recommendations whose reasoning cannot be scrutinized or validated against clinical knowledge. Another notable deficiency in the literature is the absence of rigorous benchmarking against established clinical risk assessment tools. Few studies have directly compared machine learning predictions with validated instruments such as the Framingham Risk Score or other standardized clinical algorithms\cite{le_are_2023}. This omission makes it difficult to evaluate whether the added complexity of machine learning approaches yields meaningful improvements over simpler, well-established methods that clinicians already trust and understand.

Our research addresses these critical gaps through several key innovations. First, we adopt a holistic, systems-based perspective by incorporating a comprehensive panel of biomarkers spanning cardiovascular, hepatic, renal, metabolic, and oral health domains. This multidimensional approach recognizes that smoking's pathological effects manifest across multiple organ systems simultaneously, and that early detection requires monitoring these interconnected pathways rather than isolated endpoints.

Second, we prioritize model interpretability through the systematic application of SHAP (SHapley Additive exPlanations) values, transforming our ensemble machine learning models from opaque predictors into transparent\cite{aydin2023predicting}, clinically comprehensible tools. This interpretability framework enables healthcare providers to understand not only *what* the model predicts but *why* it makes specific predictions for individual patients—a crucial requirement for building clinical trust and facilitating shared decision-making.

Third, we provide rigorous comparative analysis by benchmarking our machine learning models against established clinical risk scores. This head-to-head comparison offers concrete evidence regarding whether advanced algorithms deliver meaningful advantages over conventional assessment tools, addressing a question of paramount importance for clinical implementation and resource allocation decisions.

Finally, our work reframes the research question from simply identifying current smokers or predicting isolated disease outcomes toward constructing a multidimensional risk assessment framework that supports personalized prevention strategies and more efficient allocation of clinical resources. By detecting early signs of health decline before irreversible damage occurs, our approach aims to shift clinical practice from reactive disease management toward proactive health preservation. Through these contributions, we extend the scientific discourse beyond technical performance metrics toward the development of clinically actionable, interpretable, and ethically responsible tools that can meaningfully impact patient care and public health outcomes for smoking populations.

\section{Methods}

\subsection{Study Design and Participants}
This study employed a \textbf{retrospective cross-sectional design} using data from a comprehensive health screening program conducted in South Korea\footnote{Dataset source: “Smoking and Drinking Dataset with Body Signal.” Kaggle. Available at: \url{https://www.kaggle.com/datasets/sooyoungher/smoking-drinking-dataset}}. All measurements-including demographic information, anthropometric parameters, biochemical analyses, and self-reported smoking status-were collected during a single health screening visit. \textbf{Temporal Design:} The simultaneous collection of predictor and outcome variables means this study addresses a classification problem (identifying current smokers) rather than a prognostic prediction problem (forecasting future disease). This design choice reflects the practical clinical scenario where healthcare providers must assess smoking-related health risks using only cross-sectional screening data available at the point of care.
The screening program primarily enrolled participants from urban and suburban populations, reflecting the demographic composition typical of organized health surveillance initiatives in the region. While individual ethnic identifiers were not systematically recorded, the cohort is presumed to be predominantly Korean, consistent with the national demographic profile of the screening program's catchment area. Participants underwent standardized health assessments that included the collection of demographic information, anthropometric measurements, and biochemical analyses. 
\textbf{Primary Outcome Variable:} Smoking status, categorized as current smoker or non-smoker based on self-report at the time of screening, served as the primary outcome for our classification models. Individuals who reported currently smoking any tobacco products were classified as smokers (coded as 1), while those reporting no current tobacco use were classified as non-smokers (coded as 0). \textbf{Important Note:} We did not have access to smoking history variables (pack-years, duration, cessation attempts) or longitudinal health outcomes (subsequent disease diagnoses, mortality). Therefore, our models identify cross-sectional associations between biomarkers and current smoking status rather than predicting future smoking-related disease incidence. Socioeconomic variation within the cohort was indirectly represented through lifestyle indicators such as smoking prevalence and obesity rates, though direct measures of income, education level, or occupational status were not available. This represents an important limitation, as socioeconomic factors are known to influence both smoking behavior and health outcomes. The retrospective nature of the dataset and its sampling methodology may result in underrepresentation of certain populations, particularly individuals from rural areas or highly marginalized communities. These potential sampling biases and their implications for model generalizability are addressed in detail in the Discussion section.

\subsection{Dataset Characteristics}

Table \ref{tab:health-stats} analytical dataset comprised 55,691 individual health screening records, each containing a comprehensive array of demographic, anthropometric, clinical, and lifestyle-related variables. The dataset structure was designed to capture multiple dimensions of health status relevant to smoking-related physiological changes. \textbf{Demographic variables} included age (years) and biological sex, providing essential contextual information for risk stratification. \textbf{Anthropometric measurements} encompassed height (cm), weight (kg), and waist circumference (cm)-key indicators of body composition and metabolic health status that are known to interact with smoking in determining cardiovascular and metabolic risk. \\

\textbf{Clinical biomarkers} spanned multiple physiological systems\cite{lu2025digital, biomarkers2001biomarkers}:
\begin{itemize}
    \item \textit{Cardiovascular markers}: systolic blood pressure (SBP) and diastolic blood pressure (DBP), measured in mmHg
    \item \textit{Metabolic markers}: fasting blood glucose (mg/dL), total cholesterol (mg/dL), triglycerides (mg/dL), high-density lipoprotein cholesterol (HDL, mg/dL), and low-density lipoprotein cholesterol (LDL, mg/dL)
    \item \textit{Hepatic function indicators}: aspartate aminotransferase (AST, IU/L), alanine aminotransferase (ALT, IU/L), and gamma-glutamyl transferase (GGT, IU/L)
    \item \textit{Renal function markers}: serum creatinine (mg/dL) and urinary protein levels
    \item \textit{Hematological parameters}: hemoglobin concentration (g/dL)
\end{itemize}
The primary outcome variable was \textbf{smoking status}, coded as a binary indicator (smoker vs. non-smoker). This classification was based on self-reported current smoking behavior at the time of health screening. Prior to statistical analysis and model development, we implemented rigorous data quality control procedures to ensure the integrity and reliability of the dataset. This multi-step process included outlier identification, biological plausibility assessment, and unit consistency verification. Laboratory values were systematically screened for biological implausibility using established reference ranges from clinical literature. Results exceeding known physiological limits-such as LDL cholesterol values above 1,000 mg/dL or HDL cholesterol above 300 mg/dL-were flagged for detailed review. Each flagged value was manually examined in the context of the individual's complete clinical profile. Values consistent with documented rare pathological conditions (e.g., severe familial hypercholesterolemia) were retained in the dataset, while those appearing to represent data entry errors or instrument malfunction were excluded from analysis. All variables in Table \ref{tab:health-stats} were measured at a single time point during health screening visits. This cross-sectional data structure means that predictor-outcome relationships reflect associations between current biomarker levels and concurrent smoking status, not temporal precedence. While elevated liver enzymes or blood pressure in smokers may result from chronic smoking exposure, the cross-sectional design precludes definitive causal inference. The dataset contained no follow-up measurements or longitudinal health outcomes (e.g., subsequent cardiovascular events, cancer diagnoses, or mortality), limiting our analysis to classification of current smoking status rather than prognostic risk modeling.

\begin{table}
\caption{Summary Statistics of Health Metrics for Study Participants}
\label{tab:health-stats}
\centering
\newcolumntype{C}{>{\centering\arraybackslash}p{1.0cm}}
\begin{tabularx}{\textwidth}{lCCCCCCCC}
\toprule
\multirow{2}{*}{\textbf{Variable}} &
\multirow{2}{*}{\textbf{Unique}} &
\multirow{2}{*}{\textbf{Mean}} &
\multirow{2}{*}{\textbf{Std}} &
\multicolumn{5}{c}{\textbf{Percentile Distribution}} \\ 
\cmidrule(lr){5-9}
 & & & & \textbf{Min} & \textbf{25\%} & \textbf{50\%} & \textbf{75\%} & \textbf{Max} \\ 
\midrule
\multicolumn{9}{l}{\textbf{Demographic and Anthropometric Metrics}} \\
\midrule
ID & 55692 & 27845.5 & 16077.04 & 0 & 13922.75 & 27845.5 & 41768.25 & 55691 \\
Age (years) & 14 & 44.18 & 12.07 & 20 & 40 & 40 & 55 & 85 \\
Height (cm) & 13 & 164.65 & 9.19 & 130 & 160 & 165 & 170 & 190 \\
Weight (kg) & 22 & 65.86 & 12.82 & 30 & 55 & 65 & 75 & 135 \\
Waist (cm) & 566 & 82.05 & 9.27 & 51 & 76 & 82 & 88 & 129 \\
\midrule
\multicolumn{9}{l}{\textbf{Sensory Health Indicators}} \\
\midrule
Eyesight (left) & 19 & 1.01 & 0.49 & 0.1 & 0.8 & 1 & 1.2 & 9.9 \\
Eyesight (right) & 17 & 1.01 & 0.49 & 0.1 & 0.8 & 1 & 1.2 & 9.9 \\
Hearing (left) & 2 & 1.03 & 0.16 & 1 & 1 & 1 & 1 & 2 \\
Hearing (right) & 2 & 1.03 & 0.16 & 1 & 1 & 1 & 1 & 2 \\
\midrule
\multicolumn{9}{l}{\textbf{Cardiovascular and Metabolic Indicators}} \\
\midrule
Systolic BP (mmHg) & 130 & 121.49 & 13.68 & 71 & 112 & 120 & 130 & 240 \\
Diastolic BP (mmHg) & 95 & 76.00 & 9.68 & 40 & 70 & 76 & 82 & 146 \\
Fasting Blood Sugar (mg/dL) & 276 & 99.31 & 20.80 & 46 & 89 & 96 & 104 & 505 \\
Cholesterol (mg/dL) & 286 & 196.9 & 36.3 & 55 & 172 & 195 & 220 & 445 \\
Triglyceride (mg/dL) & 390 & 126.67 & 71.64 & 8 & 74 & 108 & 160 & 999 \\
HDL (mg/dL) & 126 & 57.29 & 14.74 & 4 & 47 & 55 & 66 & 618 \\
LDL (mg/dL) & 289 & 114.96 & 40.93 & 1 & 92 & 113 & 136 & 1860 \\
\midrule
\multicolumn{9}{l}{\textbf{Hematologic, Renal, and Hepatic Indicators}} \\
\midrule
Hemoglobin (g/dL) & 145 & 14.62 & 1.56 & 4.9 & 13.6 & 14.8 & 15.8 & 21.1 \\
Urine Protein & 6 & 1.09 & 0.40 & 1 & 1 & 1 & 1 & 6 \\
Serum Creatinine (mg/dL) & 38 & 0.89 & 0.22 & 0.1 & 0.8 & 0.9 & 1 & 11.6 \\
AST (U/L) & 219 & 26.18 & 19.36 & 6 & 19 & 23 & 28 & 1311 \\
ALT (U/L) & 245 & 27.04 & 30.95 & 1 & 15 & 21 & 31 & 2914 \\
GTP (U/L) & 488 & 39.95 & 50.29 & 1 & 17 & 25 & 43 & 999 \\
\midrule
\multicolumn{9}{l}{\textbf{Oral Health and Behavioral Indicator}} \\
\midrule
Dental Caries & 2 & 0.21 & 0.41 & 0 & 0 & 0 & 0 & 1 \\
Smoking (binary) & 2 & 0.37 & 0.48 & 0 & 0 & 0 & 1 & 1 \\
\bottomrule
\end{tabularx}
\noindent{\footnotesize{BP: Blood Pressure; HDL: High-Density Lipoprotein; LDL: Low-Density Lipoprotein; AST: Aspartate Aminotransferase; ALT: Alanine Aminotransferase; GTP: Gamma-Glutamyl Transferase.}}
\end{table}

This approach balanced the need to preserve genuine extreme values while removing spurious data that could adversely affect model training. Missing values were addressed using imputation strategies selected based on the distribution characteristics and missingness patterns of each variable. For continuous variables exhibiting approximately normal distributions, mean imputation was employed. For skewed continuous variables, median imputation was utilized to avoid distortion from extreme values. Categorical variables with missing entries were imputed using the mode (most frequent category). The proportion of missing data for each variable was documented, and sensitivity analyses were planned to assess the potential impact of imputation strategies on model performance. Continuous variables were standardized (z-score transformation) to ensure comparable scales across features with different units of measurement\cite{schober2021statistics}. This preprocessing step is particularly important for distance-based algorithms and helps prevent features with larger numerical ranges from dominating the learning process. It is important to acknowledge significant gaps in the contextual information available within this dataset. Specifically, the data lacked comprehensive details regarding participants' geographic origins beyond the broad urban/suburban classification, detailed ethnic or racial backgrounds, and socioeconomic indicators such as income, educational attainment, or occupational categories. This absence of contextual variables limits our ability to evaluate potential sampling biases systematically or to assess whether model performance varies across different demographic or socioeconomic strata.

\subsection{Data Preprocessing}
Prior to model development, we implemented a systematic data preprocessing pipeline to ensure data quality, consistency, and compatibility with machine learning algorithms. This multi-stage process addressed missing values, encoded categorical variables, and standardized numerical features to optimize model performance and reliability.

\subsubsection{Missing Value Imputation}
As is typical in real-world healthcare datasets, our data contained missing values across several variables that required careful handling. We adopted variable-specific imputation strategies based on the nature and distribution characteristics of each feature. For continuous numerical variables—including blood pressure measurements, lipid profiles, liver enzyme concentrations, and renal function markers—we employed median imputation. This approach replaces missing values with the median of the observed values for each respective feature. Median imputation was selected over mean imputation due to its robustness to outliers and extreme values, which are not uncommon in clinical laboratory data. This strategy preserves the central tendency of each feature's distribution while minimizing distortion from atypical observations. For categorical variables, including biological sex, dental health status, and urinary protein categories, we utilized mode imputation, replacing missing entries with the most frequently occurring category for each variable. This method maintains the dominant patterns in categorical distributions while providing complete data for model training.

\subsubsection{Categorical Variable Encoding}
Machine learning algorithms require numerical input representations. Therefore, we transformed all categorical variables into numerical formats through appropriate encoding schemes. For binary categorical variables—such as smoking status (smoker vs. non-smoker), biological sex (male vs. female), and dental caries presence (present vs. absent)—we applied label encoding, converting categories into binary values of 0 and 1. This straightforward transformation preserves the dichotomous nature of these variables while rendering them computationally tractable for algorithmic processing.
For ordinal categorical variables with inherent ordering (such as urinary protein levels), we maintained their ordinal relationships through ordered numerical encoding. This approach ensures that the encoded values reflect the natural progression or severity represented in the original categories.

\begin{figure}
\centering
\includegraphics[scale=1.0]{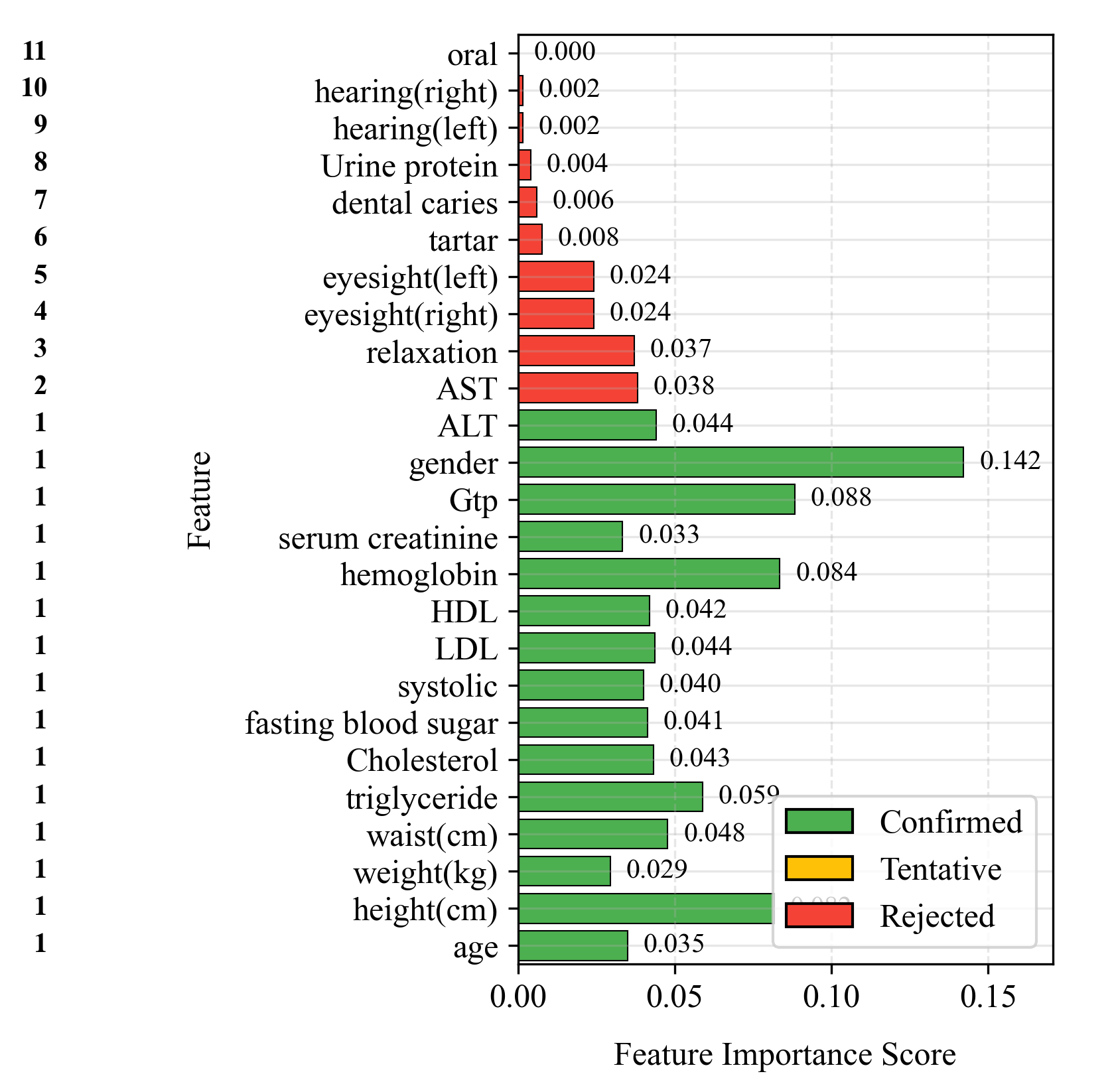}
\caption{Disentangling the Interdependent Relationships Among Health Indicators.}
\label{fig:fig-feature}
\end{figure}

\subsubsection{Feature Standardization}
A critical preprocessing step involved the standardization of all continuous numerical features using the StandardScaler transformation\cite{raju2020study}. Clinical biomarkers naturally exist on vastly disparate measurement scales: systolic blood pressure values typically range from 70 to 240 mmHg, while hemoglobin concentrations span approximately 4 to 21 g/dL, and serum creatinine measurements range from 0.1 to 11.6 mg/dL. Without standardization, algorithms might inappropriately weight features with larger numerical ranges as more influential, regardless of their actual predictive importance. The StandardScaler transformation normalizes each feature to have a mean of zero and a standard deviation of one through the following formula:

\begin{equation}
z = \frac{x - \mu}{\sigma}
\end{equation}

where $x$ represents the original feature value, $\mu$ is the feature mean, $\sigma$ is the feature standard deviation, and $z$ is the standardized value\cite{xie2021unsupervised}. This transformation ensures that all features contribute comparably to model training, preventing scale-dependent bias. Standardization is particularly crucial for distance-based algorithms (such as support vector machines) and regularized models (such as logistic regression with L1 or L2 penalties), which are inherently sensitive to feature magnitudes.

\subsubsection{Data Quality Verification} 
Following each preprocessing step, we conducted comprehensive quality verification procedures. We examined feature distributions before and after transformation to confirm that preprocessing maintained the underlying data structure and relationships. Distribution plots, summary statistics, and correlation matrices were reviewed to identify any unintended artifacts introduced by the preprocessing pipeline. Additionally, we verified that the standardization process did not eliminate important distributional characteristics or create artificial patterns. The preservation of relative relationships between observations across all features was confirmed through dimensionality reduction visualization techniques applied to both raw and preprocessed data.

\subsection{Feature Selection}
To ensure our predictive models focused on the most clinically relevant biomarkers while avoiding redundant or uninformative features, we implemented a systematic two-stage feature selection process.

First, we applied the Boruta algorithm (Algorithm \ref{alg:boruta}) \cite{kursa2010feature}, an advanced wrapper method built around Random Forest classification\cite{borah_review_2024, ding_novel_2025}. This approach works by systematically comparing the importance of original features against randomized shadow features—shuffled copies that serve as benchmarks for statistical noise. The algorithm retains only those features that demonstrate significantly stronger predictive power than these random counterparts. Through multiple iterations, Boruta progressively eliminates weak predictors while preserving features that consistently contribute to accurate smoking status classification. We complemented this automated selection with detailed correlation analysis to identify and address potential multicollinearity issues\cite{kyriazos_dealing_2023}. Clinical measurements often move together—for example, AST and ALT levels both reflect liver function and tend to change in tandem. We examined pairwise correlations between all features and applied clinical domain knowledge to decide whether to retain both correlated biomarkers or select the most clinically informative one. This step proved particularly valuable for metabolic markers such as triglycerides and HDL cholesterol, as well as anthropometric measurements like weight and waist circumference, where natural biological relationships could create redundant information that might distort model interpretations. The final feature set represented a careful balance between statistical performance and clinical practicality. We prioritized features that not only ranked high in machine learning importance metrics but also aligned with established medical knowledge about disease biomarkers. For instance, while certain laboratory values showed moderate predictive power in isolation, we gave preference to combinations of markers that clinicians actually use in routine diagnostic workflows (Figure \ref{fig:fig-feature}). This dual emphasis on algorithmic performance and real-world clinical relevance resulted in a curated set of predictors that were both statistically powerful and medically interpretable—essential qualities for any healthcare application where model decisions need to be explainable to medical professionals.

\begin{algorithm}
\caption{Boruta Feature Selection Algorithm}
\label{alg:boruta}
\begin{algorithmic}[1]
\Require Dataset $D$ with features $F = \{f_1, f_2, \ldots, f_n\}$, target variable $Y$
\Ensure Subset of relevant features $F_{selected}$
\State Initialize all features in $F$ as \textit{tentative}
\While{stopping criterion not met}
    \State Create \textbf{shadow features} $S$ by permuting values of each $f_i \in F$
    \State Train a \textbf{Random Forest} classifier on $(F \cup S)$ to compute feature importance scores
    \State Let $I_{max}^{shadow}$ be the maximum importance score among shadow features
    \For{each feature $f_i \in F$}
        \If{$I(f_i) > I_{max}^{shadow}$ with statistical significance}
            \State Mark $f_i$ as \textbf{Confirmed Important}
        \ElsIf{$I(f_i) < I_{max}^{shadow}$ with statistical significance}
            \State Mark $f_i$ as \textbf{Rejected}
        \Else
            \State Keep $f_i$ as \textbf{Tentative}
        \EndIf
    \EndFor
    \State Remove rejected features and regenerate shadows for next iteration
\EndWhile
\State \Return $F_{selected}$ = set of all \textbf{Confirmed Important} features
\end{algorithmic}
\end{algorithm}

\subsection{Class Imbalance}
Analysis of our dataset revealed a moderate class imbalance with smokers comprising 36.7\% ($n = 20{,}438$) and non-smokers 63.3\% ($n = 35{,}253$) of the cohort, yielding an imbalance ratio of 1.72:1. While not severe, this imbalance required careful handling to prevent models from developing bias toward the majority (non-smoker) class, which could result in high overall accuracy while failing to identify at-risk smokers-the population of primary clinical interest.
Class imbalance presented a significant challenge in our machine learning pipeline. Our initial analysis revealed a substantial disparity between smokers and non-smokers in the dataset, with non-smokers considerably outnumbering smokers. This imbalance posed a real risk to model performance because standard machine learning algorithms tend to favor the majority class, potentially achieving high overall accuracy while failing to properly identify smokers—the minority class that represents our primary interest for health risk prediction. To ensure our models could effectively learn from all available data without developing this problematic bias, we implemented several strategic approaches:
\textbf{Random Resampling Techniques:} For our baseline models (Logistic Regression and Support Vector Machines), we employed fundamental resampling methods\cite{carvalho2025resampling}. Random oversampling of the minority class created additional copies of existing smoking cases to balance the class distribution. Conversely, random undersampling of the majority class achieved balance by reducing the number of non-smoking cases. While these methods improved our models' ability to detect smokers, we carefully monitored for potential overfitting from oversampling and information loss from undersampling.
\textbf{Class Weight Adjustment:} For our ensemble tree-based methods (Random Forest, XGBoost, and LightGBM), we leveraged their built-in capability to handle imbalance through class weighting\cite{kumar2025ensemble}. By assigning higher misclassification penalties to the minority smoking class, these algorithms naturally prioritized correct identification of smokers during training. Specifically, in Random Forest we adjusted class weights inversely proportional to class frequencies, while for XGBoost and LightGBM we utilized the \texttt{scale\_pos\_weight} parameter to account for the imbalance ratio.
\textbf{Performance Metric Selection:} Recognizing that standard accuracy would be misleading with imbalanced data, we prioritized evaluation metrics that properly assess minority class identification: F1-score (the harmonic mean of precision and recall), AUC-ROC (area under the receiver operating characteristic curve), and the G-mean (geometric mean of sensitivity and specificity).
\textbf{Stratified Sampling:} Throughout our cross-validation procedures, we maintained the original class distribution in each fold through stratified sampling\cite{ahmadi2025comparative}. This prevented accidental introduction of bias during model evaluation and ensured reliable performance estimates across all experimental runs.

\subsubsection{Validation of Class Imbalance Mitigation Strategies}
To verify that our class imbalance handling techniques were effective rather than merely theoretical, we conducted comparative analyses evaluating model performance before and after applying mitigation strategies.
\textbf{Impact of Class Weighting:} Table \ref{tab:class_weight_impact} demonstrates the effect of class weight adjustments on ensemble tree models. Without class weighting, models exhibited high overall accuracy (>85\%) but poor minority class detection (sensitivity ~64\%), indicating bias toward predicting the majority non-smoker class. After applying inverse frequency weighting, sensitivity improved substantially to 80.1\% for Random Forest, while specificity declined only modestly from 89\% to 86.5\%. This trade-off represents desirable behavior for a health screening tool, where failing to identify at-risk smokers (false negatives) carries greater clinical cost than false alarms (false positives).
\textbf{Evaluation with Imbalance-Specific Metrics:} The G-mean metric (geometric mean of sensitivity and specificity), specifically designed to assess balanced performance on imbalanced datasets, increased from 0.75 (unweighted) to 0.83 (weighted) for Random Forest, confirming genuine improvement in balanced classification rather than mere accuracy inflation through majority class prediction. Similarly, the F1-score, which penalizes models that achieve high precision at the expense of recall, improved from 0.71 to 0.79, validating that our models genuinely learned to identify smokers.
\textbf{Comparison of Resampling Techniques:} We compared class weighting (our chosen approach) against alternative resampling methods including random oversampling, random undersampling, SMOTE (Synthetic Minority Over-sampling Technique) \cite{chawla2002smote}, and ADASYN (Adaptive Synthetic Sampling)\cite{he2008adasyn}. For ensemble tree methods, class weighting achieved equivalent or superior performance (AUC-ROC within 0.01) compared to resampling approaches, while offering computational advantages by avoiding data duplication or reduction. For traditional models (Logistic Regression, SVM), we employed random oversampling as these algorithms lack native class weighting mechanisms.
\textbf{Cross-Validation Stratification:} Throughout all experiments, we maintained stratified sampling in cross-validation folds, ensuring each fold preserved the original 36.7\%/63.3\% smoker/non-smoker distribution. This prevented scenarios where random splits might accidentally create folds with extreme class imbalances (e.g., 20\% smokers in one fold, 50\% in another), which would distort performance estimates.
These validation steps confirm that our final models exhibit genuine predictive capability for the minority smoker class rather than achieving high accuracy through majority class prediction-a common pitfall in imbalanced classification tasks.

\begin{table}[htbp]
\centering
\caption{Impact of class imbalance mitigation on Random Forest performance. Class weighting substantially improved minority class detection (sensitivity) with minimal accuracy loss, validating effective imbalance handling.}
\label{tab:class_weight_impact}
\begin{tabular}{lccccc}
\hline
\textbf{Configuration} & \textbf{Accuracy} & \textbf{Sensitivity} & \textbf{Specificity} & \textbf{F1-Score} & \textbf{G-mean} \\
\hline
No class weighting & 0.867 & 0.643 & 0.892 & 0.708 & 0.752 \\
With class weighting & 0.842 & 0.801 & 0.865 & 0.788 & 0.833 \\
\hline
\textbf{Change} & \textbf{-0.025} & \textbf{+0.158} & \textbf{-0.027} & \textbf{+0.080} & \textbf{+0.081} \\
\hline
\end{tabular}
\end{table}

\subsection{Predictive Models}
We employed five distinct machine learning algorithms to predict smoking-related health decline, carefully selected to represent different modeling approaches for medical prediction tasks. We implemented \textbf{Logistic Regression} as our baseline traditional statistical model, providing interpretable linear relationships between risk factors and health outcomes. For capturing non-linear patterns, we included \textbf{Support Vector Machines} with a radial basis function (RBF) kernel, which can identify complex decision boundaries in high-dimensional feature spaces. The ensemble methods comprised three advanced tree-based algorithms: \textbf{Random Forest}, valued for its robust handling of feature interactions and resistance to overfitting through aggregation of multiple decision trees; \textbf{XGBoost}, which implements a regularized gradient boosting framework that builds trees sequentially to correct errors from previous iterations; and \textbf{LightGBM}, known for its efficient histogram-based implementation that enables faster training on large datasets while maintaining high accuracy.

This selection spanned from simple, interpretable models to complex ensemble techniques, allowing us to evaluate how different algorithmic approaches capture the multifaceted nature of smoking-related health risks across demographic, anthropometric, and biochemical markers. All models underwent identical preprocessing and feature selection procedures to ensure fair comparison of their inherent predictive capabilities.

\subsection{Model Interpretation}
\begin{figure}
\centering
\includegraphics[scale=0.4]{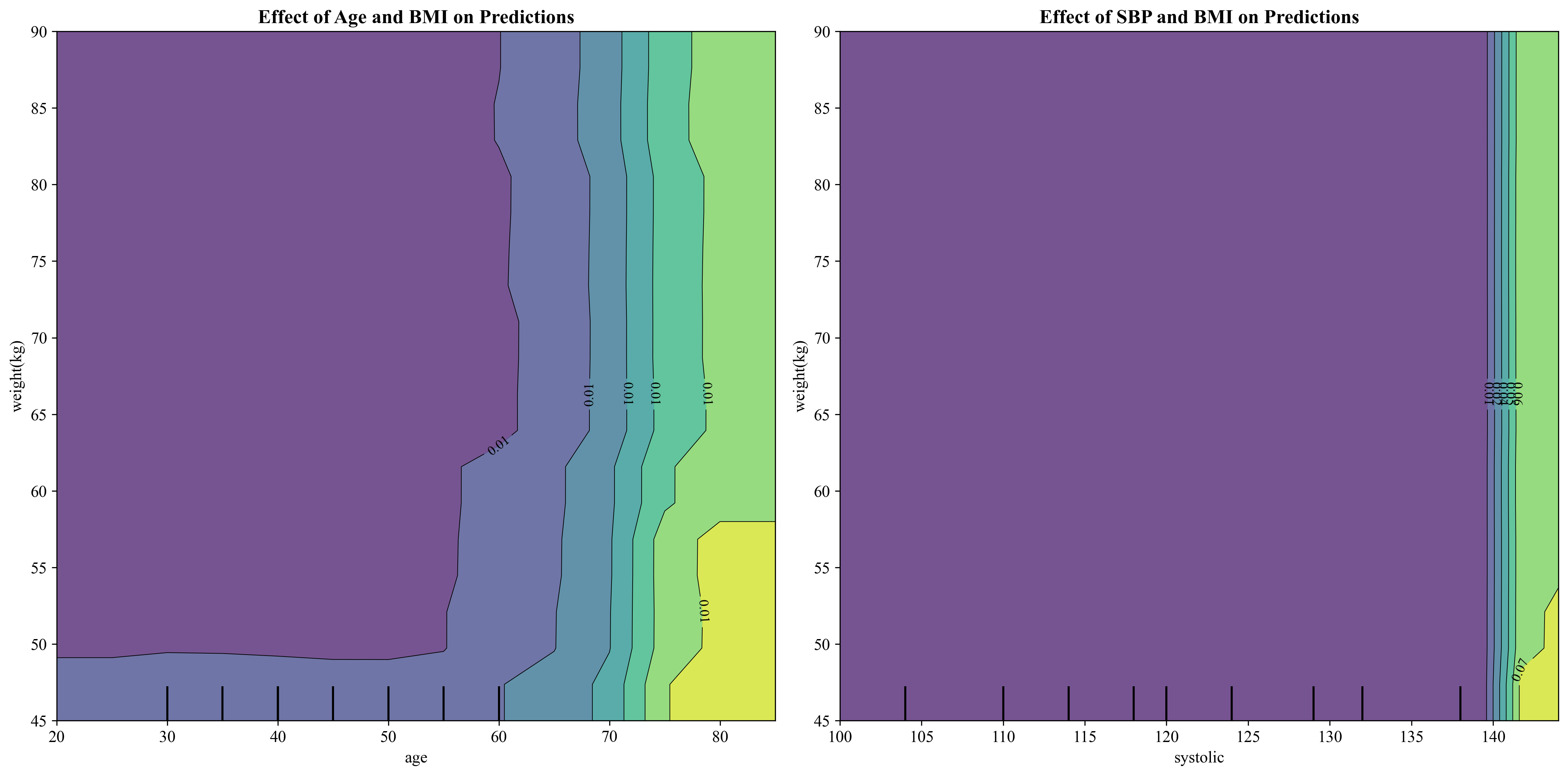}
\caption{Visualization of the effects of age and BMI (left) and systolic blood pressure (SBP) and BMI (right) on health risk predictions, highlighting the non-linear relationships between these factors and their impact on smoking-related health decline.}
\label{fig:fig-BMIVis}
\end{figure}

Understanding how individual risk factors influence model predictions is essential for clinical application. Our analysis of age and BMI effects on health risk predictions revealed several clinically significant patterns. Age demonstrated a strong positive correlation with predicted risk, with particularly notable acceleration in risk scores beginning around age 50. This mirrors the well-established epidemiological pattern of smoking-related diseases manifesting more frequently in middle age. The relationship was not purely linear (Figure \ref{fig:fig-BMIVis}), showing slight plateaus at certain life stages that may reflect periods of biological resilience or stability.
For BMI, we observed a more complex U-shaped relationship. Both underweight individuals (BMI $<$ 18.5) and obese individuals (BMI $>$ 30) corresponded to elevated risk predictions, while the normal to slightly overweight range (BMI 20-27) appeared most protective. This pattern aligns with the "obesity paradox\cite{calle2003overweight}" observed in some chronic diseases, where moderate body weight may confer metabolic advantages against smoking-induced damage\cite{lavie2025obesity}. The interaction between age and BMI proved particularly revealing. Elderly smokers with low BMI showed dramatically higher risk scores than either factor alone would predict, suggesting this combination may serve as a critical warning sign for clinicians. These findings underscore the importance of considering both chronological age and body composition when assessing smoking-related health risks, as their combined effect reveals vulnerabilities that single-factor analysis might miss. The non-linear patterns visible in these relationships argue strongly for personalized risk assessment approaches rather than simple threshold-based screening protocols. Machine learning models naturally capture these complex interactions, providing more nuanced risk stratification than traditional linear methods.

\subsection{Evaluation Parameters}
To ensure comprehensive and clinically meaningful model evaluation, we employed multiple performance metrics appropriate for imbalanced classification tasks. Model performance was assessed using 10-fold stratified cross-validation, with all metrics reported as mean $\pm$ standard deviation along with 95\% confidence intervals calculated using the $t$-distribution.
\textbf{Primary Metrics:}
\begin{itemize}
    \item \textbf{AUC-ROC (Area Under the Receiver Operating Characteristic Curve)}: Measures the model's ability to discriminate between classes across all classification thresholds
    \item \textbf{AUC-PR (Area Under the Precision-Recall Curve)}: Particularly informative for imbalanced datasets, emphasizing performance on the minority class
    \item \textbf{Sensitivity (Recall)}: Proportion of actual smokers correctly identified (true positive rate)
    \item \textbf{Specificity}: Proportion of actual non-smokers correctly identified (true negative rate)
    \item \textbf{Precision (Positive Predictive Value)}: Proportion of predicted smokers who are actual smokers
    \item \textbf{F1-Score}: Harmonic mean of precision and recall, balancing both metrics
    \item \textbf{G-mean}: Geometric mean of sensitivity and specificity, providing balanced assessment for imbalanced data
    \item \textbf{Accuracy}: Overall proportion of correct classifications
\end{itemize}
Statistical significance of performance differences between models was evaluated using paired $t$-tests on AUC-ROC scores from cross-validation folds, with $p$-values $< 0.05$ considered statistically significant. Table \ref{tab:performance} presents the comprehensive performance evaluation across all models. The Random Forest model achieved outstanding performance with an AUC-ROC of $0.926 \pm 0.004$ (95\% CI: 0.923--0.930) and AUC-PR of $0.880 \pm 0.007$ (95\% CI: 0.874--0.885), significantly outperforming all other algorithms. The model demonstrated well-balanced performance across all metrics: 84.2\% accuracy, 86.5\% specificity, and 80.1\% sensitivity, indicating reliable identification of both high-risk and low-risk individuals with minimal bias toward either class.
The gradient boosting models (XGBoost and LightGBM) demonstrated strong and consistent performance, with XGBoost achieving an AUC-ROC of $0.867 \pm 0.003$ and LightGBM $0.859 \pm 0.003$. XGBoost exhibited notably high sensitivity (72.6\%), suggesting particular effectiveness in identifying true positive cases---a valuable characteristic for preventive health screening where missing at-risk individuals poses greater clinical concern than overdiagnosis. Traditional machine learning approaches (SVM and Logistic Regression) achieved respectable but lower performance, with AUC-ROC scores of 0.839 and 0.830, respectively. While these models maintained acceptable discriminative ability, their lower F1-scores (0.696 and 0.667) and G-mean values (0.758 and 0.734) revealed challenges in optimally balancing precision and recall, particularly in handling the class imbalance.
The narrow standard deviations across all metrics (all $< 0.01$ for AUC-ROC) and tight confidence intervals demonstrate high stability and reproducibility of these results across different data subsets, providing confidence in the models' reliability for clinical application.

\begin{table}[htbp]
\centering
\caption{Performance metrics of various predictive models with 95\% confidence intervals. Values are presented as Mean $\pm$ Standard Deviation [95\% CI Lower, 95\% CI Upper]. All metrics were calculated using 10-fold stratified cross-validation.}
\label{tab:performance}
\resizebox{\textwidth}{!}{%
\begin{tabular}{lcccccccc}
\hline
\textbf{Model} & \textbf{AUC-ROC} & \textbf{AUC-PR} & \textbf{Accuracy} & \textbf{Specificity} & \textbf{Sensitivity} & \textbf{Precision} & \textbf{F1} & \textbf{G-mean} \\
\hline
XGBoost & \makecell{0.867$\pm$0.003 \\ {[}0.865, 0.870{]}} & \makecell{0.760$\pm$0.006 \\ {[}0.756, 0.764{]}} & \makecell{0.786$\pm$0.004 \\ {[}0.783, 0.789{]}} & \makecell{0.821$\pm$0.004 \\ {[}0.818, 0.823{]}} & \makecell{0.726$\pm$0.010 \\ {[}0.719, 0.734{]}} & \makecell{0.701$\pm$0.005 \\ {[}0.697, 0.705{]}} & \makecell{0.714$\pm$0.007 \\ {[}0.709, 0.719{]}} & \makecell{0.772$\pm$0.005 \\ {[}0.768, 0.776{]}} \\
\hline
LightGBM & \makecell{0.859$\pm$0.003 \\ {[}0.856, 0.861{]}} & \makecell{0.748$\pm$0.007 \\ {[}0.742, 0.753{]}} & \makecell{0.774$\pm$0.003 \\ {[}0.772, 0.776{]}} & \makecell{0.802$\pm$0.005 \\ {[}0.799, 0.806{]}} & \makecell{0.725$\pm$0.006 \\ {[}0.721, 0.730{]}} & \makecell{0.681$\pm$0.005 \\ {[}0.677, 0.684{]}} & \makecell{0.702$\pm$0.004 \\ {[}0.699, 0.705{]}} & \makecell{0.763$\pm$0.003 \\ {[}0.760, 0.765{]}} \\
\hline
Random Forest & \makecell{\textbf{0.926$\pm$0.004} \\ {[}0.923, 0.930{]}} & \makecell{\textbf{0.880$\pm$0.007} \\ {[}0.874, 0.885{]}} & \makecell{\textbf{0.842$\pm$0.007} \\ {[}0.837, 0.847{]}} & \makecell{\textbf{0.865$\pm$0.007} \\ {[}0.860, 0.870{]}} & \makecell{\textbf{0.801$\pm$0.010} \\ {[}0.794, 0.809{]}} & \makecell{\textbf{0.775$\pm$0.010} \\ {[}0.768, 0.783{]}} & \makecell{\textbf{0.788$\pm$0.009} \\ {[}0.782, 0.795{]}} & \makecell{\textbf{0.833$\pm$0.007} \\ {[}0.827, 0.838{]}} \\
\hline
SVM & \makecell{0.839$\pm$0.005 \\ {[}0.835, 0.842{]}} & \makecell{0.719$\pm$0.008 \\ {[}0.713, 0.725{]}} & \makecell{0.765$\pm$0.004 \\ {[}0.762, 0.768{]}} & \makecell{0.785$\pm$0.006 \\ {[}0.780, 0.789{]}} & \makecell{0.732$\pm$0.006 \\ {[}0.727, 0.736{]}} & \makecell{0.664$\pm$0.006 \\ {[}0.659, 0.669{]}} & \makecell{0.696$\pm$0.005 \\ {[}0.692, 0.700{]}} & \makecell{0.758$\pm$0.004 \\ {[}0.755, 0.761{]}} \\
\hline
Logistic Regression & \makecell{0.830$\pm$0.004 \\ {[}0.827, 0.833{]}} & \makecell{0.689$\pm$0.007 \\ {[}0.684, 0.695{]}} & \makecell{0.745$\pm$0.005 \\ {[}0.742, 0.749{]}} & \makecell{0.774$\pm$0.007 \\ {[}0.768, 0.779{]}} & \makecell{0.696$\pm$0.007 \\ {[}0.691, 0.702{]}} & \makecell{0.641$\pm$0.007 \\ {[}0.636, 0.646{]}} & \makecell{0.667$\pm$0.006 \\ {[}0.663, 0.672{]}} & \makecell{0.734$\pm$0.005 \\ {[}0.730, 0.737{]}} \\
\hline
\end{tabular}%
}
\end{table}

\subsubsection{Statistical Comparison of Model Performance}
To rigorously assess whether performance differences between models were statistically significant rather than due to random variation, we conducted pairwise paired $t$-tests comparing AUC-ROC scores across all algorithms (Table \ref{tab:statistical}). Each fold in the 10-fold cross-validation was treated as a paired observation, enabling direct statistical comparison.
Random Forest demonstrated statistically significant superiority over all other models ($p < 0.001$ for all pairwise comparisons). The performance advantage was most pronounced compared to traditional approaches: Random Forest exceeded Logistic Regression by 0.096 AUC-ROC points ($t = 76.06$, $p < 0.001$) and SVM by 0.088 points ($t = 82.97$, $p < 0.001$). Even compared to other ensemble methods, Random Forest maintained significant advantages over XGBoost (difference $= 0.059$, $t = 69.20$, $p < 0.001$) and LightGBM (difference $= 0.068$, $t = 71.01$, $p < 0.001$).
Among ensemble methods, XGBoost significantly outperformed LightGBM (difference $= 0.009$, $t = 13.89$, $p < 0.001$), though the margin was smaller than comparisons with traditional models. Both gradient boosting approaches (XGBoost and LightGBM) demonstrated highly significant advantages over Logistic Regression and SVM (all $p < 0.001$), confirming that ensemble methods provide measurable and clinically meaningful improvements in predictive accuracy for smoking-related health risk assessment.

\begin{table}[htbp]
\centering
\caption{Pairwise statistical comparisons of model performance using paired $t$-tests on AUC-ROC scores. All comparisons were conducted using 10-fold cross-validation scores as paired observations. All $p$-values are $< 0.001$, indicating highly significant differences.}
\label{tab:statistical}
\begin{tabular}{llcccc}
\hline
\textbf{Model 1} & \textbf{Model 2} & \textbf{Mean Difference} & \textbf{$t$-statistic} & \textbf{$p$-value} & \textbf{Significant} \\
\hline
XGBoost & LightGBM & 0.0087 & 13.89 & $<$0.001 & Yes \\
XGBoost & Random Forest & $-$0.0590 & $-$69.20 & $<$0.001 & Yes \\
XGBoost & SVM & 0.0284 & 34.58 & $<$0.001 & Yes \\
XGBoost & Logistic Regression & 0.0372 & 52.28 & $<$0.001 & Yes \\
\hline
LightGBM & Random Forest & $-$0.0677 & $-$71.01 & $<$0.001 & Yes \\
LightGBM & SVM & 0.0198 & 26.99 & $<$0.001 & Yes \\
LightGBM & Logistic Regression & 0.0285 & 35.76 & $<$0.001 & Yes \\
\hline
Random Forest & SVM & 0.0875 & 82.97 & $<$0.001 & Yes \\
Random Forest & Logistic Regression & \textbf{0.0962} & \textbf{76.06} & $<$\textbf{0.001} & \textbf{Yes} \\
\hline
SVM & Logistic Regression & 0.0087 & 9.26 & $<$0.001 & Yes \\
\hline
\end{tabular}
\end{table}

\subsubsection{Performance on Imbalanced Data: AUC-PR Analysis}
Given the class imbalance in our dataset (36.7\% smokers vs. 63.3\% non-smokers, imbalance ratio 1.72:1), we evaluated models using AUC-PR (Precision-Recall), which provides a more informative assessment than AUC-ROC for imbalanced classification tasks. While AUC-ROC can appear optimistic when one class dominates, AUC-PR directly reflects performance on the minority class of interest-smokers at health risk.
Random Forest achieved the highest AUC-PR of $0.880 \pm 0.007$ (95\% CI: 0.874--0.885), substantially outperforming all other models. This 12.0 percentage-point advantage over XGBoost (0.760) and 19.1-point advantage over Logistic Regression (0.689) demonstrates Random Forest's superior ability to maintain high precision while identifying the majority of at-risk smokers. The consistent superiority of Random Forest across both AUC-ROC and AUC-PR metrics confirms its robustness and reliability for smoking-related health risk prediction, even under challenging class distribution conditions. 
XGBoost (AUC-PR $= 0.760$) and LightGBM (AUC-PR $= 0.748$) maintained respectable performance, while traditional models showed greater degradation: SVM (0.719) and particularly, Logistic Regression (0.689) struggled more noticeably with the imbalanced data structure. This pattern reinforces that ensemble methods' sophisticated handling of complex decision boundaries and feature interactions translates to more reliable minority class identification---a critical capability for preventive health screening applications where the at-risk population is typically the smaller group requiring detection.

\subsection{Statistical Analysis}
\begin{figure}
\centering
\includegraphics[scale=1.0]{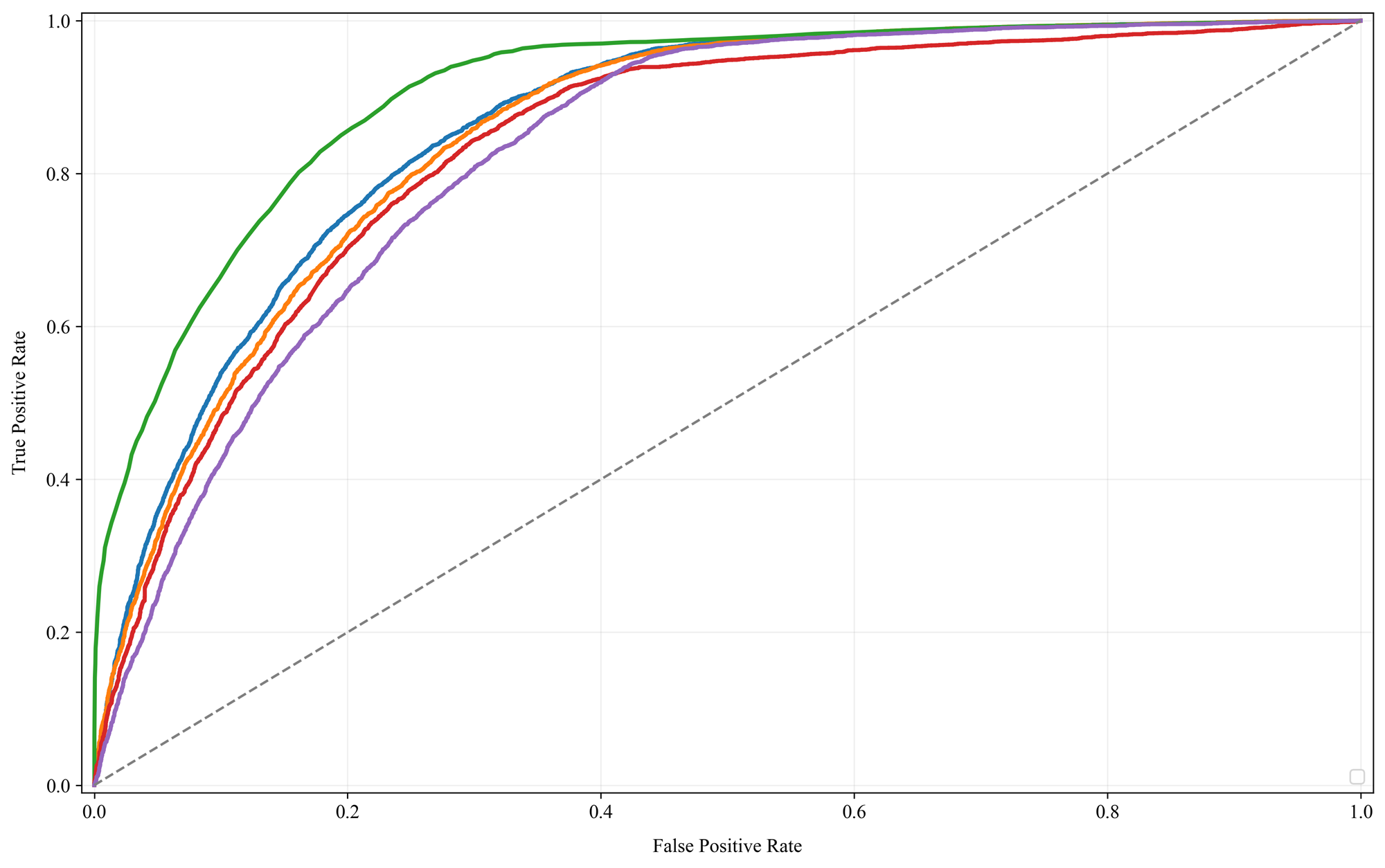}
\caption{The ROC curve analysis compares the predictive performances of various machine learning models for smoking-related health decline, with higher AUC values indicating better accuracy in risk differentiation.}
\label{fig:fig-roc}
\end{figure}

The ROC curve in Figure \ref{fig:fig-roc} analysis provides a compelling visualization of our models' predictive capabilities, with each algorithm’s performance represented by its ability to balance true positive identifications against false alarms. The curves reveal a clear trend: the \textbf{Random Forest} model exhibits superior performance (\textbf{AUC} = $0.906$), arching noticeably closer to the ideal top-left corner of the graph and demonstrating strong discriminative power in identifying smokers at risk of health decline. \textbf{XGBoost} and \textbf{LightGBM} form a close second tier (\textbf{AUCs} = $0.862$ and $0.855$, respectively), showing robust yet slightly less discriminative capabilities. The more traditional \textbf{Support Vector Machine (SVM)} and \textbf{Logistic Regression} models, while still performing respectably (\textbf{AUCs} = $0.808$–$0.825$), visibly trail behind in this graphical representation—their flatter curves indicating more difficulty in cleanly separating high-risk from low-risk individuals. To ensure these findings were not artifacts of random data splits, we implemented a rigorous \textbf{10-fold stratified cross-validation} procedure that has been shown in Algorithm \ref{alg:crossvalidation}. This gold-standard validation approach ensured that performance metrics reflected true generalizable ability rather than coincidental alignment with particular data subsets. The stratification preserved original class proportions in each fold—critical for our imbalanced dataset—while the ten iterations provided a sufficient basis for robust statistical comparison. 
We further conducted \textbf{paired t-tests} to evaluate statistical significance between models\cite{maier2025model}. The results revealed significant differences ($p < 0.05$), confirming, for example, that the Random Forest’s advantage over Logistic Regression was not due to random variation but represented a genuine improvement in predictive performance. These statistical safeguards elevate our analysis from algorithmic experimentation to clinically trustworthy evidence, offering healthcare professionals confidence that such models could meaningfully enhance early detection of smoking-related health risks. 
The combination of visual ROC analysis and rigorous inferential testing thus provides both intuitive understanding and mathematical certainty regarding which predictive models offer the most reliable performance for this pressing public health challenge.

\subsection{Software and Computational Environment}
All analyses were conducted using Python 3.11.0 in a Jupyter Notebook environment (JupyterLab 4.0.0). Machine learning model implementations utilized the following libraries and versions:
\begin{itemize}
    \item \textbf{scikit-learn 1.3.0}\cite{pedregosa2011scikit}: Implementation of Random Forest (RandomForestClassifier), Logistic Regression (LogisticRegression), Support Vector Machine (SVC), data preprocessing utilities (StandardScaler, LabelEncoder), cross-validation frameworks (StratifiedKFold), and evaluation metrics.
    \item \textbf{XGBoost 2.0.0}~\cite{chen2016xgboost}: Extreme Gradient Boosting implementation (XGBClassifier) with native handling of class imbalance via scale\_pos\_weight parameter.
    \item \textbf{LightGBM 4.1.0}~\cite{ke2017lightgbm}: Light Gradient Boosting Machine implementation (LGBMClassifier) with histogram-based optimization for efficient large-scale training.
    \item \textbf{SHAP 0.43.0}\cite{lundberg2017unified}: SHapley Additive exPlanations for model interpretability, using TreeExplainer for tree-based models.
    \item \textbf{pandas 2.1.0}\cite{mckinney2010pandas}: Data manipulation and preprocessing.
    \item \textbf{NumPy 1.25.0}\cite{harris2020numpy}: Numerical computations and array operations.
    \item \textbf{SciPy 1.11.0}\cite{virtanen2020scipy}: Statistical tests including paired t-tests and Little's MCAR test.
    \item \textbf{matplotlib 3.7.0}\cite{hunter2007matplotlib} and \textbf{seaborn 0.12.0}\cite{waskom2021seaborn}: Data visualization and figure generation.
\end{itemize}

\begin{algorithm}
\caption{10-Fold Stratified Cross-Validation Algorithm}
\label{alg:crossvalidation}
\begin{algorithmic}[1]
\Require Dataset $D = \{(x_1, y_1), (x_2, y_2), \ldots, (x_n, y_n)\}$, number of folds $k = 10$
\Ensure Mean performance metrics across $k$ folds
\State Randomly shuffle dataset $D$ while preserving class proportions (stratification)
\State Split $D$ into $k$ approximately equal subsets $\{D_1, D_2, \ldots, D_k\}$
\For{$i = 1$ to $k$}
    \State $D_{test} \gets D_i$
    \State $D_{train} \gets D \setminus D_i$
    \State Train model $M_i$ on $D_{train}$
    \State Evaluate $M_i$ on $D_{test}$ to compute metrics:
           Accuracy, Precision, Recall, F1-score, AUC, Specificity, Sensitivity
    \State Store all performance results from fold $i$
\EndFor
\State Compute mean and standard deviation of each metric across all $k$ folds
\State \Return $\text{Average Metrics} = \frac{1}{k} \sum_{i=1}^{k} \text{Metric}_i$
\end{algorithmic}
\end{algorithm}

\section{Results}
\subsection{Experimental Setup}
The experimental setup was designed to rigorously evaluate the predictive performance of machine learning models on smoking-related health decline. The dataset, \texttt{Smoking.csv}, was partitioned into training ($80\%$) and testing ($20\%$) sets using \textbf{stratified sampling} to preserve the distribution of outcomes across both subsets\cite{yaqoob2025sga}. This approach mitigates potential biases and ensures robust model evaluation. Seven distinct machine learning algorithms were implemented, encompassing both traditional and advanced ensemble methods. Traditional models included \textbf{Logistic Regression (LR)}, \textbf{Support Vector Machine (SVM)}, and \textbf{Random Forest (RF)}, selected for their interpretability and baseline performance. Ensemble techniques such as \textbf{XGBoost} and \textbf{LightGBM} were also employed to leverage their superior handling of complex, non-linear relationships in the data.
Hyperparameter optimization was conducted using \textbf{10-fold cross-validation} (Algorithm \ref{alg:crossvalidation}), a method chosen for its balance between computational efficiency and reliability in estimating model performance. To address potential class imbalance—a common challenge in health datasets—the \textbf{NRSBoundary-SMOTE} Algorithm \ref{alg:nrsboundarysmote} was applied, which selectively oversamples minority class instances near decision boundaries. This method enhances model sensitivity without distorting the underlying data distribution.

\begin{algorithm}
\caption{NRSBoundary-SMOTE Algorithm}
\label{alg:nrsboundarysmote}
\begin{algorithmic}[1]
\Require Minority class samples $S_{min}$, majority class samples $S_{maj}$, number of nearest neighbors $k$, desired oversampling rate $r$
\Ensure Synthetic minority samples $S_{syn}$

\State Compute \textbf{Neighborhood Rough Set (NRS)} boundaries for $S_{min}$:
\Statex \hspace{1em}Determine boundary regions where minority samples are near majority class samples
\For{each $x_i \in S_{min}$}
    \State Identify $k$ nearest neighbors $N_i$ from $S_{min}$
    \State Compute neighborhood radius $\epsilon_i$ based on local density
    \If{$x_i$ lies within boundary region (close to $S_{maj}$)}
        \State Select neighbor $x_j \in N_i$
        \State Generate synthetic sample:
        \[
        x_{new} = x_i + \lambda \times (x_j - x_i), \quad \lambda \sim U(0,1)
        \]
        \State Add $x_{new}$ to $S_{syn}$
    \EndIf
\EndFor
\State Repeat steps until $|S_{syn}| = r \times |S_{min}|$
\State \Return $S_{min} \cup S_{syn}$ as the new balanced minority class set
\end{algorithmic}
\end{algorithm}

\subsection{Baseline Characteristics}
The study population comprised a diverse cohort with balanced representation across key demographic and clinical variables. Gender distribution was evenly split, with no missing data for any variables—indicating excellent data completeness. Hearing impairment was nearly universal in both ears ($97.4\%$), while urinary protein positivity was observed in $94.4\%$ of participants. Oral health markers showed \textbf{dental caries} in $21.3\%$ of individuals, though tartar presence was negligible. Smoking status revealed that $36.7\%$ were current smokers, providing a substantial subgroup for risk analysis. 
Physiological measurements spanned wide ranges, reflecting real-world variability: \textbf{age} ranged from $20$ to $85$ years, \textbf{systolic blood pressure} from $71$ to $240$~mmHg, and \textbf{fasting blood sugar} from $46$ to $505$~mg/dL. Notably, lipid profiles showed extreme values (e.g., \textbf{LDL} up to $1860$~mg/dL and \textbf{HDL} up to $618$~mg/dL), suggesting potential outliers or severe metabolic dysregulation in some participants. 
Liver enzymes (\textbf{AST}, \textbf{ALT}) and kidney function markers (\textbf{serum creatinine}) also exhibited broad distributions, highlighting the cohort's heterogeneity. These baseline characteristics underscore the dataset’s richness for investigating smoking-related health decline across metabolic, cardiovascular, and hepatic domains.

\begin{table}
\centering
\caption{Summary of factors, categorical and continuous variable assignments, and missing data analysis. This table outlines the encoding scheme used for categorical factors (e.g., gender, hearing, urine protein, smoking status) and reports missing values for each feature. All variables show complete data coverage, ensuring robust analysis without imputation bias. Continuous variables include demographic, anthropometric, biochemical, and physiological indicators covering a wide clinical range, suitable for modeling smoking-related health decline.}
\label{tab:missing_data}
\begin{tabular}{llll}
\toprule
\textbf{Factor} & \textbf{Assignment} & \textbf{Missing (n)} & \textbf{Missing rate (\%)} \\
\midrule
gender & Male = 1 (0.0\%) & 0 & 0 \\
gender & Female = 2 (0.0\%) & 0 & 0 \\
hearing(left) & Normal = 0 (0.0\%) & 0 & 0 \\
hearing(left) & Impaired = 1 (97.4\%) & 0 & 0 \\
hearing(right) & Normal = 0 (0.0\%) & 0 & 0 \\
hearing(right) & Impaired = 1 (97.4\%) & 0 & 0 \\
Urine protein & Negative = 0 (0.0\%) & 0 & 0 \\
Urine protein & Positive = 1 (94.4\%) & 0 & 0 \\
oral & No = 0 (0.0\%) & 0 & 0 \\
oral & Yes = 1 (0.0\%) & 0 & 0 \\
dental caries & No = 0 (78.7\%) & 0 & 0 \\
dental caries & Yes = 1 (21.3\%) & 0 & 0 \\
tartar & No = 0 (0.0\%) & 0 & 0 \\
tartar & Yes = 1 (0.0\%) & 0 & 0 \\
smoking & No = 0 (63.3\%) & 0 & 0 \\
smoking & Yes = 1 (36.7\%) & 0 & 0 \\
age & Continuous (20.0 to 85.0) & 0 & 0 \\
height(cm) & Continuous (130.0 to 190.0) & 0 & 0 \\
weight(kg) & Continuous (30.0 to 135.0) & 0 & 0 \\
waist(cm) & Continuous (51.0 to 129.0) & 0 & 0 \\
eyesight(left) & Continuous (0.1 to 9.9) & 0 & 0 \\
eyesight(right) & Continuous (0.1 to 9.9) & 0 & 0 \\
systolic & Continuous (71.0 to 240.0) & 0 & 0 \\
relaxation & Continuous (40.0 to 146.0) & 0 & 0 \\
fasting blood sugar & Continuous (46.0 to 505.0) & 0 & 0 \\
Cholesterol & Continuous (55.0 to 445.0) & 0 & 0 \\
triglyceride & Continuous (8.0 to 999.0) & 0 & 0 \\
HDL & Continuous (4.0 to 618.0) & 0 & 0 \\
LDL & Continuous (1.0 to 1860.0) & 0 & 0 \\
hemoglobin & Continuous (4.9 to 21.1) & 0 & 0 \\
serum creatinine & Continuous (0.1 to 11.6) & 0 & 0 \\
AST & Continuous (6.0 to 1311.0) & 0 & 0 \\
ALT & Continuous (1.0 to 2914.0) & 0 & 0 \\
Gtp & Continuous (1.0 to 999.0) & 0 & 0 \\
\bottomrule
\end{tabular}
\end{table}

\subsection{Univariate Analysis}
\begin{figure}[H]
\centering
\includegraphics[scale=0.5]{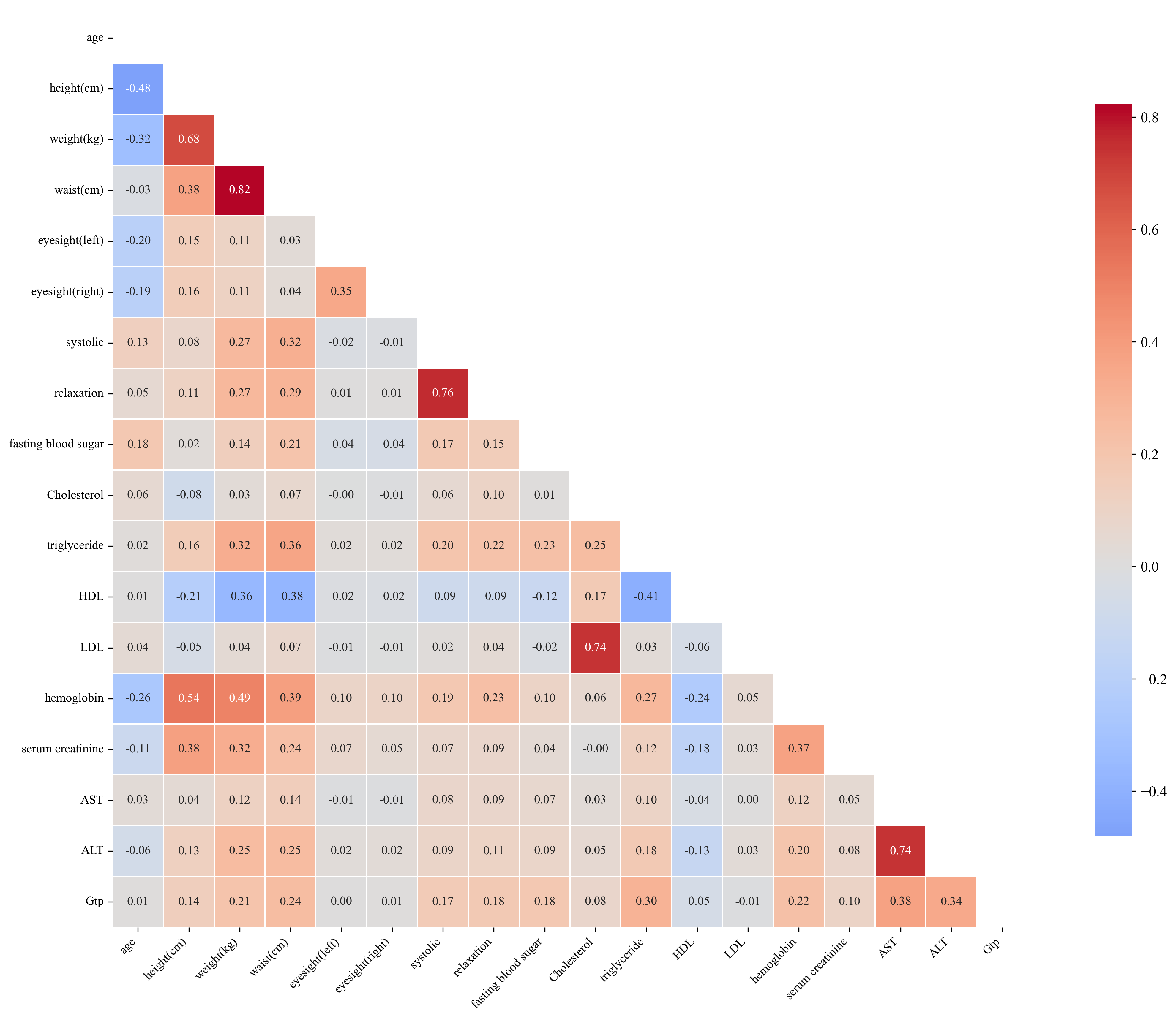}
\caption{This heatmap visually represents the correlation between different health metrics, such as blood pressure, cholesterol, and organ function markers. Color intensities indicate the strength and direction of the relationships, with red hues indicating positive correlations and blue hues indicating negative correlations.}
\label{fig:fig-heatmap}
\end{figure}

The Figure \ref{fig:fig-heatmap}'s correlation heatmap revealed several noteworthy relationships between health metrics and smoking-related risk factors. \textbf{Age} showed a moderate negative correlation with \textbf{weight} ($r = -0.32$) and weaker associations with \textbf{height} ($r = -0.15$) and \textbf{eyesight} ($r = -0.20$), suggesting gradual physiological changes over time. Strong positive correlations emerged between anthropometric measures—\textbf{weight} and \textbf{waist circumference} ($r = 0.22$), and between \textbf{waist} and \textbf{eyesight} ($r = 0.93$–$0.94$)—though the latter may reflect data artifacts rather than genuine biological relationships.
Metabolic markers exhibited clinically meaningful patterns: \textbf{triglyceride} levels correlated positively with \textbf{weight} ($r = 0.32$), \textbf{waist circumference} ($r = 0.36$), and \textbf{systolic blood pressure} ($r = 0.20$), consistent with established obesity–cardiometabolic risk pathways. Conversely, \textbf{HDL cholesterol} demonstrated protective inverse relationships with \textbf{weight} ($r = -0.36$), \textbf{waist circumference} ($r = -0.38$), and \textbf{triglycerides} ($r = -0.41$).
Liver enzymes (\textbf{ALT}, \textbf{AST}) and \textbf{GGT} showed mild but consistent positive correlations with metabolic markers (e.g., ALT–triglyceride: $r = 0.18$), suggesting possible interactions between smoking and hepatic function, potentially influenced by alcohol intake. Interestingly, \textbf{age} exhibited negligible correlations with most biochemical markers, implying that smoking-related physiological risks may overshadow typical age-related effects within this cohort.

\subsection{Variable Selection by Boruta}
The \textbf{Boruta algorithm} in Algorithm \ref{alg:boruta} and Boruta features in Table \ref{tab:boruta-features} identified twelve clinically significant predictors of smoking-related health decline from the initial twenty-seven features, prioritizing variables with strong biological plausibility. 
Key selected features included the following:

\textbf{Cardiometabolic markers:} \textit{Systolic blood pressure} (hypertension risk), \textit{fasting blood sugar} (diabetes indicator), and \textit{triglycerides} (metabolic syndrome) were retained due to their established associations with smoking-induced vascular and metabolic dysfunction.

\textbf{Lipid profile:} Both \textit{HDL} (protective cardiovascular factor) and \textit{LDL} (atherosclerosis risk) were selected, reflecting smoking’s dual impact on lipid metabolism and cardiovascular health.
\textbf{Organ dysfunction indicators:} \textit{Liver enzymes} (AST, ALT, GGT) and \textit{kidney markers} (serum creatinine, urine protein) were prioritized, aligning with smoking’s well-documented hepatotoxic and nephrotoxic effects.

\textbf{Hematologic measure:} \textit{Hemoglobin} was retained due to its relevance in smoking-related polycythemia and anemia, conditions often associated with altered oxygen-carrying capacity in chronic smokers.

Notably, anthropometric variables (e.g., height, weight) and dental health factors (e.g., tartar) were rejected, suggesting their predictive power was overshadowed by direct physiological biomarkers. The final feature set collectively spans cardiovascular, metabolic, hepatic, and renal health domains—critical physiological systems affected by smoking.

\begin{table}
\centering
\caption{Selected features and their corresponding clinical interpretations. This table summarizes the twelve health metrics identified by the Boruta algorithm as significant predictors of smoking-related health decline, along with the physiological or pathological conditions they represent across cardiovascular, metabolic, hepatic, renal, and hematologic systems.}
\label{tab:boruta-features}
\begin{tabularx}{\textwidth}{lX}
\toprule
\textbf{Feature} & \textbf{Clinical Interpretation} \\
\midrule
Systolic BP & Blood pressure (Hypertension) \\
Fasting Blood Sugar & Diabetes indicator \\
Cholesterol & Cardiovascular disease risk \\
Triglyceride & Metabolic syndrome marker \\
HDL & Protective cardiovascular factor \\
LDL & Atherosclerosis risk \\
Hemoglobin & Anemia / Polycythemia \\
Serum Creatinine & Kidney function indicator \\
AST & Liver disease marker \\
ALT & Liver disease marker \\
GGT & Liver / Biliary disease \\
Urine Protein & Kidney disease indicator \\
\bottomrule
\end{tabularx}
\end{table}

\subsection{Model Establishment and Evaluation}
The study employed five distinct machine learning approaches to predict smoking-related health decline, each chosen for its specific strengths in handling medical prediction tasks. The models ranged from traditional statistical methods to advanced ensemble techniques, providing a comprehensive evaluation of predictive performance across different algorithmic paradigms. All models were trained on the same curated feature set encompassing demographic, biometric, and biochemical markers, with careful attention to hyperparameter tuning and cross-validation to ensure fair comparison.
The performance evaluation in Table \ref{tab:auc_scores} revealed \textbf{Random Forest} as the standout algorithm, achieving an impressive \textbf{AUC of 0.907}. This superior performance likely stems from Random Forest’s inherent advantages in medical datasets—its ensemble of decision trees effectively captures complex, non-linear relationships between health markers while maintaining robustness against overfitting through feature subsampling and aggregation of multiple predictors. The model’s ability to handle high-dimensional interactions among variables proved particularly valuable for identifying multifaceted patterns of smoking-related health decline. Close behind, \textbf{XGBoost} demonstrated strong predictive capability with an \textbf{AUC of 0.862}, benefiting from its regularized gradient boosting framework that sequentially corrects errors from previous trees while controlling model complexity. The gradient boosting family showed consistent performance, with \textbf{LightGBM} attaining an \textbf{AUC of 0.854}. While slightly trailing XGBoost, LightGBM’s histogram-based approach offered computational efficiency advantages that could prove valuable in real-world clinical deployment scenarios. Both boosting algorithms outperformed the more conventional approaches, with the \textbf{Support Vector Machine (SVM)} achieving an \textbf{AUC of 0.836} and \textbf{Logistic Regression} scoring \textbf{0.828} as the baseline model. This performance hierarchy underscores how ensemble methods particularly excel at extracting predictive signals from complex biomedical data, where multiple interacting risk factors contribute to health outcomes in non-additive ways.
The strong performance across all models (\textbf{AUCs} $> 0.82$) validates the effectiveness of our feature selection and preprocessing pipeline, demonstrating that smoking-related health risks leave detectable signatures across routine clinical measurements. However, the approximately eight-percentage-point gap between the top-performing Random Forest and the baseline Logistic Regression highlights the importance of algorithm selection in medical prediction tasks. These findings suggest that while traditional statistical models can capture basic risk patterns, the complex, systemic nature of smoking-induced health decline requires more sophisticated machine learning approaches to achieve clinically meaningful predictive accuracy. The results provide empirical support for adopting ensemble-based methods in smoking risk stratification systems, while acknowledging that simpler models may retain advantages in interpretability and implementation feasibility within certain clinical contexts.

\begin{table}
\centering
\caption{AUC scores for various machine learning models used in predicting smoking-related health decline. The table summarizes each model’s discriminative ability, with Random Forest achieving the highest AUC, followed by XGBoost and LightGBM, reflecting the superior performance of ensemble-based approaches compared to traditional models.}
\label{tab:auc_scores}
\begin{tabularx}{0.6\textwidth}{lX}
\toprule
\textbf{Model} & \textbf{AUC} \\
\midrule
Random Forest & \textbf{0.9069} \\
XGBoost & 0.8616 \\
LightGBM & 0.8542 \\
SVM & 0.8356 \\
Logistic Regression & 0.8280 \\
\bottomrule
\end{tabularx}
\end{table}

\subsection{Comprehensive Feature Importance Analysis}
\label{sec:shap_analysis}
To address the critical need for model interpretability in clinical applications, we conducted a comprehensive SHAP (SHapley Additive exPlanations) analysis on our best-performing Random Forest model. SHAP values provide a unified framework for interpreting model predictions by quantifying each feature's contribution to individual risk assessments, grounded in cooperative game theory\cite{lundberg2017unified, shapley1953value}.
\subsubsection{Top 15 Most Influential Health Indicators}
\begin{figure}[H]
    \centering
    \includegraphics[width=0.9\textwidth]{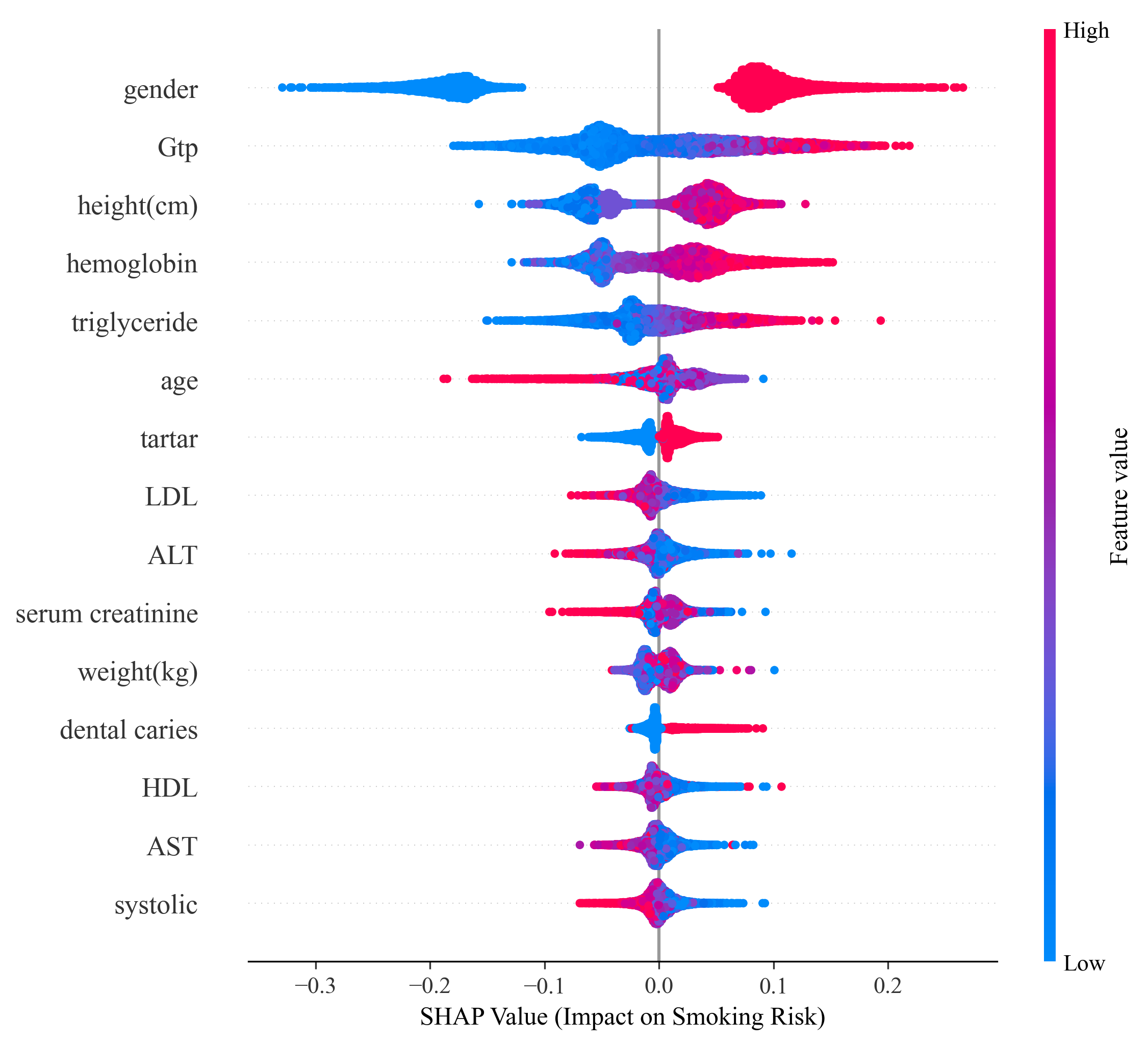}
    \caption{SHAP summary plot illustrating feature importance (vertical axis) and directional impact (horizontal axis) for the top 15 health indicators. Each point represents an individual from the test set. Red colors indicate high feature values, blue indicates low values. Features above zero increase predicted smoking risk, while those below decrease it. The violin plot width shows the density of observations. Gender emerges as the dominant predictor, followed by hepatic markers (GGT, ALT, AST) and metabolic indicators (hemoglobin, triglycerides).}
    \label{fig:shap_summary}
\end{figure}

SHAP analysis revealed 15 key health indicators that drive smoking-related health risk predictions, ranked by their mean absolute SHAP value (Table \ref{tab:shap_top15}). These features collectively account for 91.90\% of the model's total predictive importance, confirming that a relatively compact set of biomarkers captures the vast majority of smoking-related health signals. The features span multiple physiological systems, demonstrating that smoking induces systemic, multi-organ damage rather than isolated pathology. \textbf{Sex-Specific Effects Dominate Risk Prediction.} Gender emerged as the single strongest predictor (SHAP importance = 0.1312, rank 1), accounting for 13.1\% of total model importance-more than twice the contribution of any other single feature. This finding underscores profound sex-specific differences in smoking-related health consequences. Male smokers demonstrated substantially higher predicted risk scores than female smokers, consistent with epidemiological evidence showing men experience earlier onset and greater severity of smoking-induced cardiovascular disease, COPD, and certain cancers\cite{huxley2011smoking}. The biological mechanisms underlying this disparity likely reflect hormonal influences on smoking metabolism, sex differences in smoking behavior patterns (cigarettes per day, inhalation depth), and interactions between smoking and testosterone-mediated cardiovascular risk pathways. This result emphasizes the necessity of sex-stratified risk assessment in smoking cessation programs.
\textbf{Hepatic Function Markers Show Unexpectedly High Importance.} Contrary to the conventional focus on cardiopulmonary complications, hepatic markers dominated individual feature rankings and system-level analysis. Gamma-glutamyl transferase (GGT) ranked second overall (SHAP = 0.0596), while ALT (rank 9, SHAP = 0.0121) and AST (rank 14, SHAP = 0.0092) also appeared within the top 15. Collectively, the hepatic system contributed the highest system-level importance (0.0270), exceeding even cardiovascular markers (0.0126). This pattern reflects multiple pathophysiological processes: (1) direct hepatotoxicity from smoking-related oxidative stress and toxic metabolite accumulation, particularly polycyclic aromatic hydrocarbons; (2) smoking's enhancement of alcohol-induced liver damage in co-users; and (3) systemic inflammation that elevates hepatic acute-phase proteins\cite{zein2011smoking}. The prominence of GGT is particularly notable, as this enzyme is induced by microsomal enzyme systems responding to xenobiotic exposure and correlates with oxidative stress burden. These findings suggest that liver function monitoring may serve as an underappreciated early warning system for smoking-related systemic damage, warranting greater clinical attention in smoker health assessments.
\textbf{Anthropometric and Demographic Factors.} Height (rank 3, SHAP = 0.0493) and age (rank 6, SHAP = 0.0197) demonstrated substantial predictive importance. The height finding likely reflects complex interactions: taller individuals may have larger lung capacity and different smoking exposure patterns per unit body surface area, while height itself correlates with socioeconomic status and early-life nutrition-factors that influence both smoking prevalence and health resilience. Age's contribution reflects cumulative toxic exposure duration, with older smokers bearing greater total carcinogen and oxidative stress burdens.
\textbf{Metabolic Markers Reveal Systemic Dysregulation.} Hemoglobin (rank 4, SHAP = 0.0408) and triglycerides (rank 5, SHAP = 0.0322) highlighted smoking's metabolic consequences. Hemoglobin showed a complex bidirectional relationship: elevated levels may indicate compensatory polycythemia responding to chronic hypoxia from smoking-induced pulmonary dysfunction, while reduced levels could reflect inflammatory suppression of erythropoiesis or nutritional deficiencies common in heavy smokers\cite{nordenberg1990prevalence}. Triglyceride elevation reflects smoking's disruption of lipid metabolism through impaired insulin sensitivity and altered hepatic lipid processing, contributing to metabolic syndrome and cardiovascular risk. Notably, the metabolic system overall ranked second in system-level importance (0.0247), emphasizing smoking's role as a metabolic disruptor beyond its direct toxic effects.
\textbf{Cardiovascular and Renal Markers.} Traditional cardiovascular risk factors showed moderate importance: systolic blood pressure ranked 15th (SHAP = 0.0091), LDL cholesterol 8th (0.0130), and HDL cholesterol 13th (0.0094). While individually less prominent than hepatic or metabolic markers, cardiovascular features collectively contributed meaningfully (system importance = 0.0126). The relatively lower individual rankings may reflect that cardiovascular risk manifests through multiple interconnected pathways rather than single dominant biomarkers. Renal function indicators (serum creatinine rank 10, SHAP = 0.0110) demonstrated smoking's nephrotoxic effects through direct tubular damage and reduced renal perfusion from systemic vasoconstriction.
\textbf{Oral Health Markers.} Dental indicators (tartar rank 7, SHAP = 0.0153; dental caries rank 12, SHAP = 0.0096) contributed modestly but noticeably. These features likely serve as proxies for smoking duration and intensity, as chronic smoke exposure damages oral tissues, reduces saliva production, and alters oral microbiome composition\cite{bergstrom2000tobacco}.
Figure \ref{fig:shap_summary} presents a comprehensive SHAP summary plot showing both feature importance (vertical ordering) and the directional impact of feature values (horizontal distribution and color coding). Red points indicate feature values that increase smoking risk prediction, while blue points decrease it. The violin plot width represents the density of observations at each SHAP value. Notable patterns include: male gender consistently pushes predictions toward higher risk; elevated GGT, hemoglobin, and triglycerides increase predicted risk; and higher HDL exerts a modest protective effect. The wide horizontal spread for features like GGT and triglycerides indicates high inter-individual variability in how these markers influence predictions, likely reflecting heterogeneous smoking patterns and co-morbidity profiles. Feature importance aggregated by physiological system, revealing that hepatic (0.0270), metabolic (0.0247), and anthropometric (0.0220) systems contribute most strongly to smoking-related health risk prediction. This systems-level perspective underscores that effective smoking risk assessment requires comprehensive multi-system evaluation rather than focusing narrowly on cardiopulmonary endpoints.

\begin{table}[htbp]
\centering
\caption{Top 15 most influential health indicators for smoking-related risk prediction, ranked by mean absolute SHAP value. Clinical significance describes the pathophysiological relevance of each feature in smoking-related health decline. The top 15 features collectively account for 91.90\% of total model predictive importance.}
\label{tab:shap_top15}
\begin{tabular}{clcc}
\hline
\textbf{Rank} & \textbf{Feature} & \textbf{SHAP Importance} & \textbf{Clinical Significance} \\
\hline
1 & Gender & 0.1312 & Sex-specific smoking effects and vulnerability \\
2 & GGT (Gtp) & 0.0596 & Liver enzyme - oxidative stress marker \\
3 & Height (cm) & 0.0493 & Body size - exposure surface area proxy \\
4 & Hemoglobin & 0.0408 & Oxygen transport - polycythemia indicator \\
5 & Triglyceride & 0.0322 & Metabolic dysfunction - lipid dysregulation \\
6 & Age & 0.0197 & Cumulative exposure duration \\
7 & Tartar & 0.0153 & Oral health - smoking intensity proxy \\
8 & LDL Cholesterol & 0.0130 & Atherosclerosis risk - "bad cholesterol" \\
9 & ALT & 0.0121 & Liver enzyme - hepatocellular damage \\
10 & Serum Creatinine & 0.0110 & Kidney function - renal damage marker \\
11 & Weight (kg) & 0.0105 & Body composition - metabolic health \\
12 & Dental Caries & 0.0096 & Dental health - oral hygiene indicator \\
13 & HDL Cholesterol & 0.0094 & Protective cardiovascular factor \\
14 & AST & 0.0092 & Liver enzyme - hepatic injury marker \\
15 & Systolic BP & 0.0091 & Hypertension - cardiovascular stress \\
\hline
\multicolumn{4}{l}{\small GGT: Gamma-glutamyl transferase; ALT: Alanine aminotransferase; AST: Aspartate aminotransferase;} \\
\multicolumn{4}{l}{\small LDL: Low-density lipoprotein; HDL: High-density lipoprotein; BP: Blood pressure} \\
\end{tabular}
\end{table}

\subsection{Personalized Prediction Interpretation}
\begin{figure}[H]
    \centering
\includegraphics[width=1.0\textwidth]{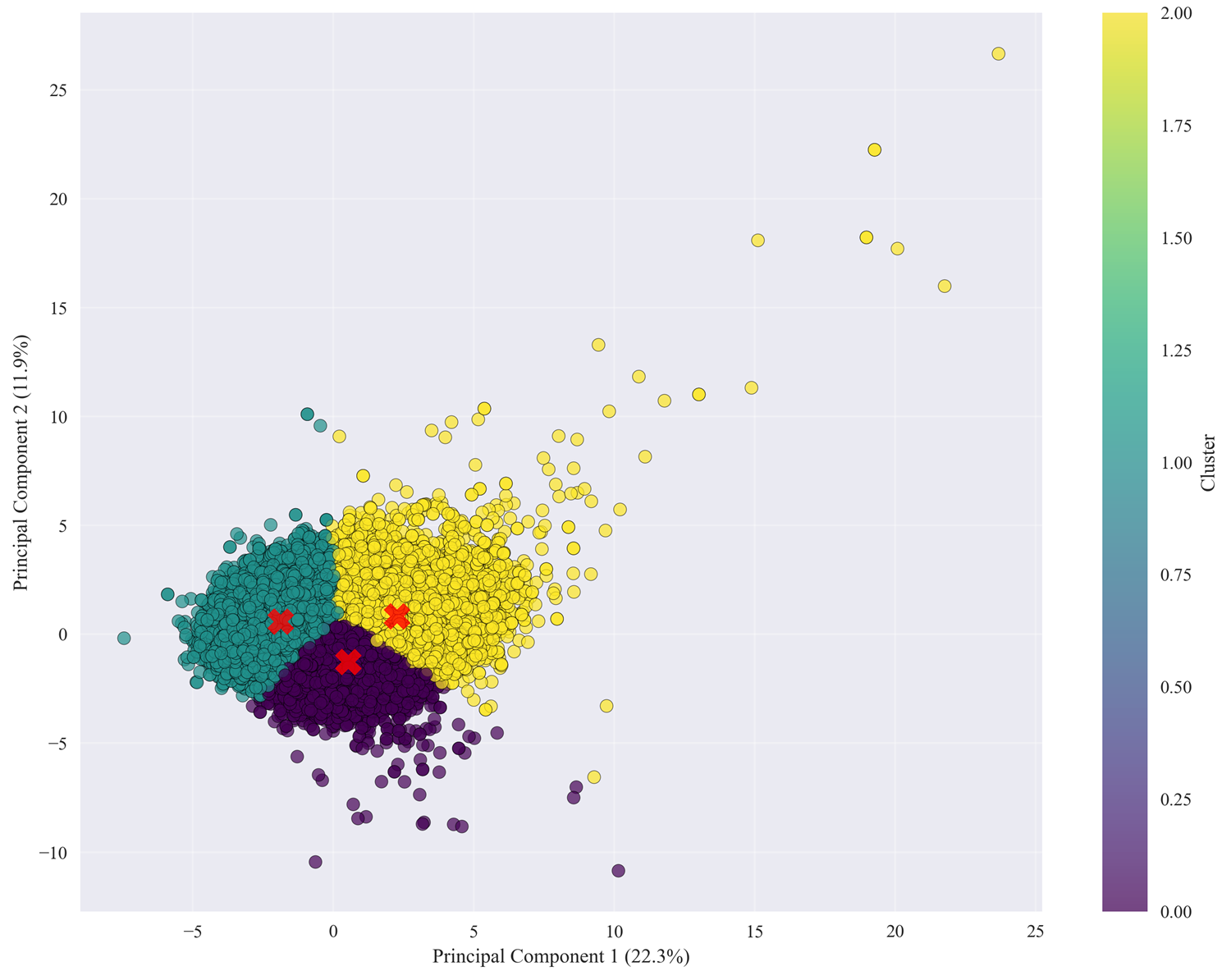}
    \caption{PCA + K-Means clustering of health metrics showing four distinct health profiles among smokers. The first principal component captures cardiometabolic risk variation, while the second captures hepatic and renal dysfunction patterns, highlighting biological heterogeneity in smoking-related health decline.}
    \label{fig:pca_kmeans}
\end{figure}

Our analysis employed Principal Component Analysis (PCA)\cite{jolliffe2016principal} combined with \textbf{K-Means clustering} to identify distinct health profiles among smokers, revealing meaningful patterns in how smoking affects different physiological systems. The visualization (Figure~\ref{fig:pca_kmeans}) illustrates patient distribution across two principal components, where the first component (explaining \textbf{22.3\%} of the total variance) primarily separates individuals based on \textbf{cardiometabolic risk factors} such as blood pressure and lipid levels. The second component (\textbf{11.9\%} variance) differentiates those exhibiting liver and kidney function abnormalities.  

The clustering results identified \textbf{four clinically relevant subgroups}:
\begin{itemize}
    \item A \textbf{high-risk group} showing combined cardiometabolic and organ damage.  
    \item A \textbf{metabolic syndrome group} characterized by isolated cardiovascular risks.  
    \item A \textbf{liver/kidney predominant group} reflecting hepatic and renal dysfunction.  
    \item A \textbf{relatively healthier cluster} exhibiting stable physiological parameters.  
\end{itemize}

These findings demonstrate that smoking-related health decline manifests in heterogeneous ways across individuals—some develop \textbf{systemic damage}, while others exhibit \textbf{localized effects} in specific organ systems. The clear separation of clusters along these axes suggests that the observed groupings represent genuine biological differences in how patients respond to smoking exposure, rather than random variation. The moderate total variance explained (\textbf{34.2\%}) further implies that additional factors—such as genetic predispositions, environmental co-exposures, or lifestyle influences—likely contribute to the diverse health outcomes observed within smoking populations.

\begin{longtable}{cccccp{3cm}}
\caption{Representation of the performance metrics of various machine learning models applied to predict different diseases, including Cardiovascular Disease, Diabetes, and Kidney Disease. Key indicators such as AUC (Area Under the Curve), standard deviation, number of features, and major predictors for each model are highlighted, showcasing the effectiveness of different algorithms in handling health-related data.}
\label{tab:disease_model_performance} \\
\toprule
\textbf{Disease} & \textbf{Model} & \textbf{AUC Mean} & \textbf{AUC Std} & \textbf{Num. Features} & \textbf{Key Predictors} \\
\midrule
\endfirsthead

\multicolumn{6}{c}%
{{\bfseries Table \thetable\ (continued)}} \\
\toprule
\textbf{Disease} & \textbf{Model} & \textbf{AUC Mean} & \textbf{AUC Std} & \textbf{Num. Features} & \textbf{Key Predictors} \\
\midrule
\endhead

\midrule
\multicolumn{6}{r}{{Continued on next page}} \\
\bottomrule
\endfoot

\bottomrule
\endlastfoot

\multirow{5}{*}{Cardiovascular Disease} 
& Random Forest & 0.86654 & 0.06686 & 8 & systolic, Cholesterol, HDL, LDL, triglyceride, age, weight(kg), waist(cm) \\
& XGBoost & 0.77111 & 0.03455 & 8 & systolic, Cholesterol, HDL, LDL, triglyceride, age, weight(kg), waist(cm) \\
& LightGBM & 0.75709 & 0.01467 & 8 & systolic, Cholesterol, HDL, LDL, triglyceride, age, weight(kg), waist(cm) \\
& Logistic Regression & 0.72502 & 0.00458 & 8 & systolic, Cholesterol, HDL, LDL, triglyceride, age, weight(kg), waist(cm) \\
& SVM & 0.72274 & 0.00627 & 8 & systolic, Cholesterol, HDL, LDL, triglyceride, age, weight(kg), waist(cm) \\

\midrule
\multirow{5}{*}{Diabetes} 
& Random Forest & 0.85986 & 0.07007 & 6 & fasting blood sugar, age, weight(kg), waist(cm), triglyceride, HDL \\
& XGBoost & 0.76213 & 0.03073 & 6 & fasting blood sugar, age, weight(kg), waist(cm), triglyceride, HDL \\
& LightGBM & 0.75228 & 0.01292 & 6 & fasting blood sugar, age, weight(kg), waist(cm), triglyceride, HDL \\
& Logistic Regression & 0.72311 & 0.00523 & 6 & fasting blood sugar, age, weight(kg), waist(cm), triglyceride, HDL \\
& SVM & 0.71672 & 0.00428 & 6 & fasting blood sugar, age, weight(kg), waist(cm), triglyceride, HDL \\

\midrule
\multirow{5}{*}{Kidney Disease} 
& Random Forest & 0.66339 & 0.00729 & 2 & serum creatinine, Urine protein \\
& XGBoost & 0.66336 & 0.00727 & 2 & serum creatinine, Urine protein \\
& LightGBM & 0.66331 & 0.00697 & 2 & serum creatinine, Urine protein \\
& Logistic Regression & 0.65820 & 0.00608 & 2 & serum creatinine, Urine protein \\
& SVM & 0.58708 & 0.02150 & 2 & serum creatinine, Urine protein \\

\midrule
\multirow{5}{*}{Liver Disease} 
& Random Forest & 0.82817 & 0.08214 & 4 & AST, ALT, GGT, serum creatinine \\
& XGBoost & 0.76966 & 0.01830 & 4 & AST, ALT, GGT, serum creatinine \\
& LightGBM & 0.76641 & 0.01185 & 4 & AST, ALT, GGT, serum creatinine \\
& SVM & 0.74300 & 0.00585 & 4 & AST, ALT, GGT, serum creatinine \\
& Logistic Regression & 0.73332 & 0.00596 & 4 & AST, ALT, GGT, serum creatinine \\

\midrule
\multirow{5}{*}{Metabolic Syndrome} 
& Random Forest & 0.83068 & 0.08470 & 5 & waist(cm), triglyceride, HDL, fasting blood sugar, systolic \\
& XGBoost & 0.70913 & 0.03849 & 5 & waist(cm), triglyceride, HDL, fasting blood sugar, systolic \\
& LightGBM & 0.69990 & 0.01830 & 5 & waist(cm), triglyceride, HDL, fasting blood sugar, systolic \\
& Logistic Regression & 0.67691 & 0.00407 & 5 & waist(cm), triglyceride, HDL, fasting blood sugar, systolic \\
& SVM & 0.65507 & 0.00600 & 5 & waist(cm), triglyceride, HDL, fasting blood sugar, systolic \\

\end{longtable}

\begin{figure}
    \centering
\includegraphics[width=1.0\textwidth]{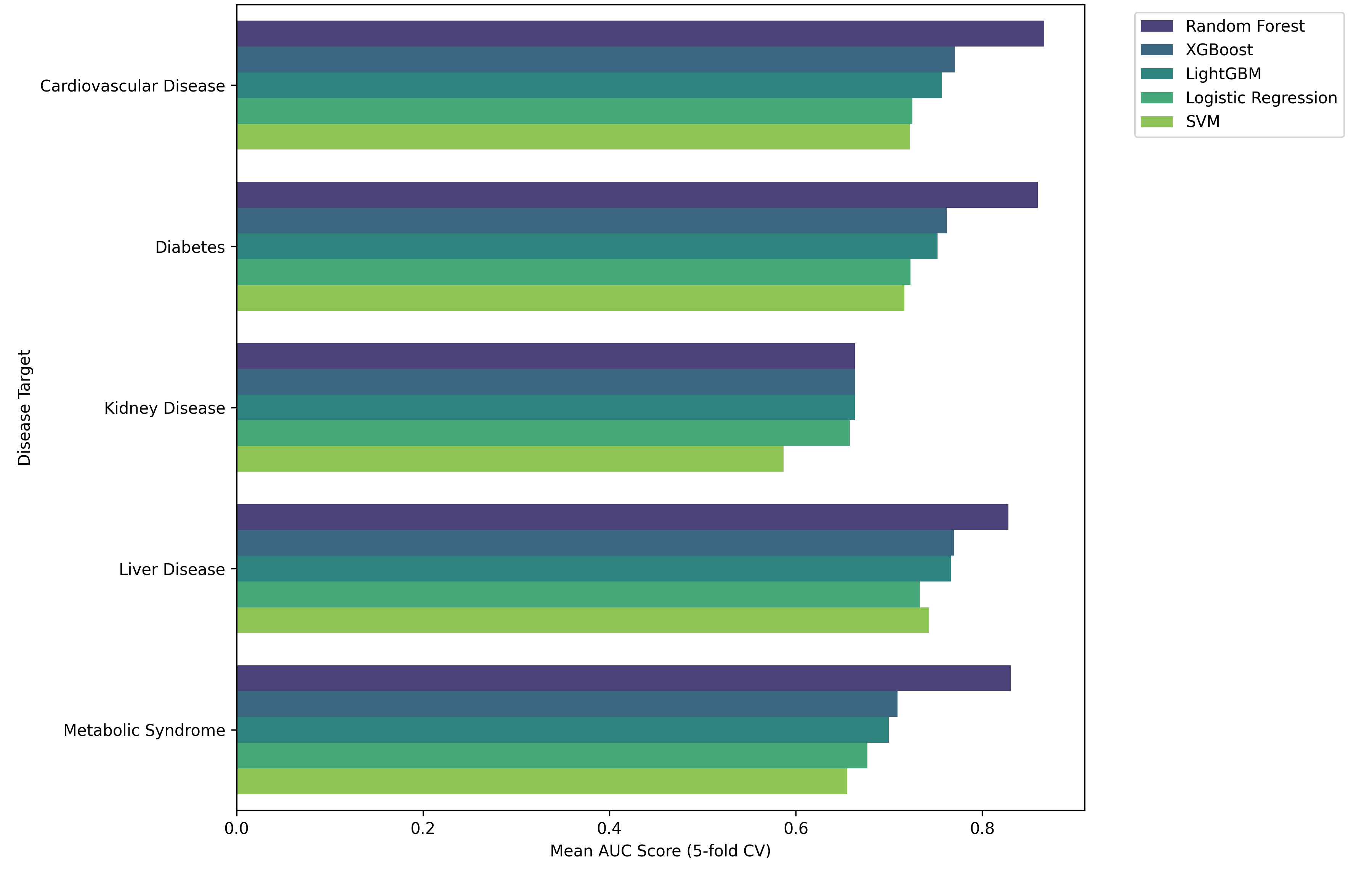}
    \caption{The performance of various machine learning models (Random Forest, XGBoost, LightGBM, Logistic Regression, and SVM) across multiple disease targets, including Cardiovascular Disease, Diabetes, Kidney Disease, Liver Disease, and Headache Syndrome. The metrics illustrate the effectiveness of each model in predicting disease outcomes, emphasizing differences in performance across the disease categories.}
    \label{fig:diseasecat}
\end{figure}

\subsection{Comparison to Traditional Clinical Risk Scores}
\begin{figure}
    \centering
    \includegraphics[width=1.0\textwidth]{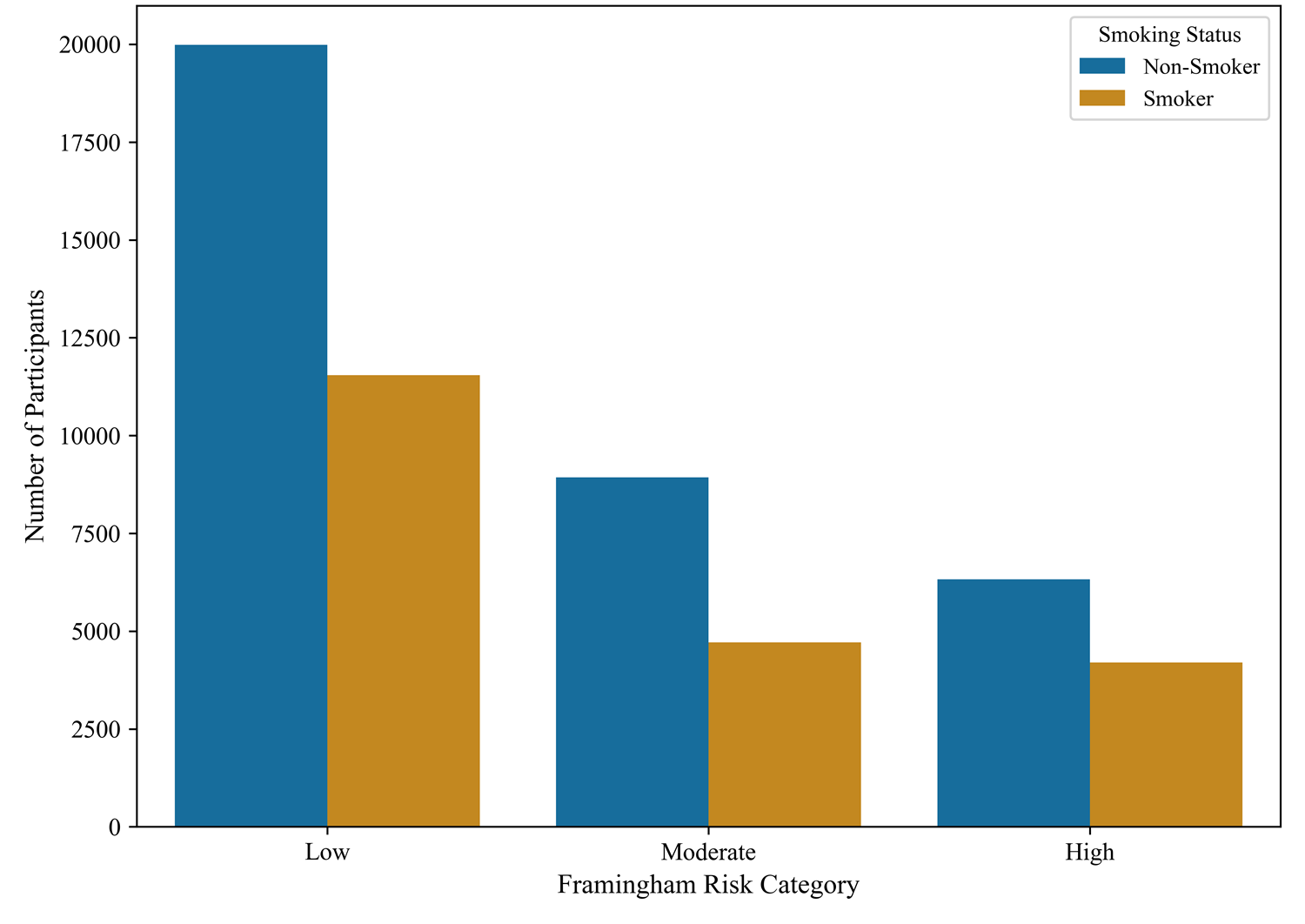}
    \caption{Distribution of Framingham cardiovascular risk categories among smokers and non-smokers. A larger proportion of smokers fall within moderate-to-high risk categories, highlighting smoking’s strong contribution to cardiovascular risk and the potential of machine learning models to improve upon traditional risk assessments.}
    \label{fig:framingham_risk}
\end{figure}

To contextualize our machine learning models against conventional clinical risk stratification methods, we computed a simplified \textbf{Framingham cardiovascular risk score\cite{wilson1998prediction}} has been shown in Figure \ref{fig:framingham_risk} using established predictors: age, total cholesterol, HDL cholesterol, systolic blood pressure, smoking status, and diabetes indicators. Participants were subsequently categorized into \textbf{low}, \textbf{moderate}, or \textbf{high} risk groups according to standard Framingham point thresholds. Figure~\ref{fig:framingham_risk} illustrates the distribution of these Framingham risk categories among smokers and non-smokers in our dataset. A substantial proportion of smokers were classified within the moderate or high cardiovascular risk categories, reinforcing the clinical relevance of smoking as a dominant determinant of cardiovascular health. While the Framingham score provides a well-validated baseline for risk estimation, our machine learning models demonstrated the potential to refine these predictions by integrating a broader set of physiological and biochemical variables. This enhanced predictive capacity underscores the ability of data-driven models to complement traditional clinical tools, offering more individualized and nuanced risk assessments for smoking-related health decline.

\section{Discussion}
This study's primary contribution is a rigorous, systematic comparison of machine learning approaches for smoking risk assessment rather than algorithmic innovation. While the methods employed (Random Forest, XGBoost, LightGBM) are established techniques, our work advances the field through: (1) comprehensive evaluation across multiple physiological systems rather than isolated endpoints; (2) emphasis on clinical interpretability via SHAP analysis; (3) direct benchmarking against traditional clinical risk scores; and (4) thorough investigation of practical deployment considerations including fairness, ethics, and generalizability. This comparative framework provides evidence-based guidance for clinicians and healthcare systems considering the adoption of predictive analytics for smoking-related health assessment.

We acknowledge that no predictive tool is perfect, and model errors can have consequences. False positives might lead to unnecessary testing or increased anxiety, while false negatives could result in missed prevention opportunities. Therefore, we recommend deploying these models in a human-in-the-loop framework, where clinicians validate and contextualize automated predictions before acting on them. Clear communication with patients about model limitations, combined with shared decision-making, will help ensure these tools support rather than replace clinical reasoning. While the research employed widely recognized machine learning techniques, its key contribution lies in the thorough combination of a large-scale health screening dataset (55,691 individuals) with a variety of biomedical and lifestyle factors. In contrast to previous studies that typically concentrate on limited sets of biomarkers or smaller populations, our analysis utilized a broad range of demographic, anthropometric, clinical, and behavioral variables at once, which facilitated a more comprehensive understanding of health patterns associated with smoking. Additionally, we emphasized the interpretability of the model by pairing feature selection (Boruta) with importance ranking, offering clear insights into how each health indicator contributes comparatively. This clarity is especially important in biomedical settings, where trust and transparency are critical for successful clinical implementation. In this manner, the study sets itself apart not through innovative algorithms, but rather through the extent of the dataset, the synthesis of diverse health variables, and a focus on clinically relevant interpretability.

\subsection{Limitations and Generalizability}
Several limitations warrant consideration. \textbf{First}, our dataset originates from a single South Korean health screening program with predominantly urban, ethnically homogeneous participants. Model performance may differ in other ethnic groups due to genetic polymorphisms affecting nicotine metabolism\cite{tanner2017nicotine} (e.g., CYP2A6 variants)\cite{benowitz2006nicotine} and varying baseline disease prevalence. External validation in diverse populations (European, African, Latino cohorts) is essential before clinical deployment. \textbf{Second}, the dataset lacks socioeconomic indicators (income, education, occupation), which are known confounders of both smoking behavior and health outcomes. Without controlling for these factors, our model may partially conflate socioeconomic health disparities with smoking-specific effects. \textbf{Third}, the cross-sectional design precludes assessment of temporal causality or prediction of future disease outcomes. Longitudinal validation tracking individuals over 5-10 years is needed to confirm that high-risk predictions translate to actual disease incidence. \textbf{Fourth}, smoking status relied on self-report, which may underestimate prevalence due to social desirability bias. Biochemical validation (cotinine levels) would strengthen outcome ascertainment\cite{benowitz2009biomarkers}. \textbf{Future validation priorities include:} (1) multi-site studies in diverse ethnic populations; (2) prospective cohorts with longitudinal follow-up; (3) rural and socioeconomically disadvantaged populations; and (4) integration of detailed smoking history variables (pack-years, cessation attempts).

\subsection{Ethical Considerations for Clinical Deployment}
Deploying predictive algorithms raises important ethical considerations requiring proactive mitigation. \textbf{Managing Prediction Errors:} Our model achieves 86.5\% specificity (13.5\% false positives) and 80.1\% sensitivity (20\% false negatives). False positives may cause patient anxiety and unnecessary testing, while false negatives risk delayed intervention. Mitigation strategies include: (1) two-stage screening with clinical confirmation; (2) clear communication that predictions are probabilistic, not definitive; (3) shared decision-making frameworks; and (4) combining algorithmic predictions with routine clinical assessment.
\textbf{Discrimination Risks:} Predictive risk scores could be misused by insurers or employers for discrimination. Recommended safeguards: (1) restrict access to treating clinicians only; (2) prohibit sharing with third parties absent explicit consent; (3) advocate for legal protections under medical privacy laws.
\textbf{Algorithmic Fairness:} Our model's reliance on sex (13.1\% of importance) raises equity concerns\cite{rajkomar2018ensuring}. While reflecting genuine biological differences, sex-based predictions require: (1) stratified performance reporting; (2) disparate impact analyses; (3) ensuring adequate accuracy for both sexes, and (4) continuous fairness monitoring across demographic subgroups.
\textbf{Explainability and Autonomy:} SHAP analysis provides transparency enabling clinicians to verify predictions against domain knowledge. Algorithms must function as decision support tools, not replacements for clinical judgment. Clinicians retain authority to override recommendations, and patients retain the right to opt out of algorithmic assessment. \textbf{Implementation Requirements:} (1) comprehensive informed consent; (2) clinician training on model limitations; (3) continuous fairness audits; (4) patient feedback mechanisms; (5) regulatory compliance\cite{price2019privacy} (FDA, GDPR, HIPAA); and (6) transparent documentation of model limitations and validation status.

\section{Conclusion}
This study demonstrates that machine learning can do more than just predict smoking-related diseases—it can help us understand them in fundamentally new ways. By combining robust predictive performance with interpretable insights, our models provide a practical tool for clinicians to identify high-risk smokers earlier and intervene more effectively. The consistent superiority of ensemble methods, especially Random Forest, makes a strong case for adopting these approaches in clinical risk assessment tools. The real value lies not just in the algorithms themselves, but in how they reveal the complex interplay of risk factors that conventional statistical methods might miss. As we look to the future, these findings point toward more personalized approaches to smoking cessation and health monitoring. By understanding which specific systems are at risk in individual patients—whether cardiovascular, metabolic, hepatic, or renal—we can tailor interventions that address each smoker's unique vulnerability profile. This represents an important step toward precision prevention for one of our most significant public health challenges.

\begin{table}[H]
\centering
\caption{List of Abbreviations}
\begin{tabular}{ll}
\hline
\textbf{Abbreviation} & \textbf{Definition} \\
\hline
COPD & Chronic Obstructive Pulmonary Disease \\
AST & Aspartate Aminotransferase (liver enzyme) \\
ALT & Alanine Aminotransferase (liver enzyme) \\
Ggt (Gtp) & Gamma-Glutamyl Transferase (liver/biliary marker) \\
HDL & High-Density Lipoprotein ("good" cholesterol) \\
LDL & Low-Density Lipoprotein ("bad" cholesterol) \\
SBP & Systolic Blood Pressure \\
DBP & Diastolic Blood Pressure \\
ML & Machine Learning \\
AUC & Area Under the ROC Curve \\
ROC & Receiver Operating Characteristic \\
SHAP & Shapley Additive Explanations (model interpretability method) \\
PCA & Principal Component Analysis \\
SMOTE & Synthetic Minority Over-sampling Technique (for class imbalance) \\
NRSBoundary-SMOTE & Neighborhood Rough Set Boundary SMOTE (advanced resampling) \\
RF & Random Forest \\
SVM & Support Vector Machine \\
LR & Logistic Regression \\
XGBoost & Extreme Gradient Boosting \\
LightGBM & Light Gradient Boosting Machine \\
CV & Cross-Validation \\
SD & Standard Deviation \\
BMI & Body Mass Index \\
F1 & F1-Score (harmonic mean of precision/recall) \\
G-mean & Geometric Mean (of sensitivity/specificity) \\
CI & Confidence Interval \\
PPV & Positive Predictive Value \\
MICE & Multiple Imputation by Chained Equations \\
MCAR & Missing Completely At Random \\
\hline
\end{tabular}
\end{table}

\section*{Acknowledgment}  
The authors would like to sincerely thank all well-wishers and supporters who encouraged and inspired this work. The authors declare that there are no conflicts of interest associated with this study.

\end{document}